\documentclass{article}
\usepackage{iclr2027_conference,times}

\usepackage{amsmath,amsfonts,bm}

\def\eqref#1{equation~\ref{#1}}
\def\1{\bm{1}}

\DeclareMathAlphabet{\mathsfit}{\encodingdefault}{\sfdefault}{m}{sl}
\SetMathAlphabet{\mathsfit}{bold}{\encodingdefault}{\sfdefault}{bx}{n}

\usepackage{hyperref}
\usepackage{xcolor}
\usepackage{url}
\usepackage{cleveref}

\title{World Observer: \\Joint Actor-Observer Generation for \\Persistent World Modeling}

\author{%
\textbf{Hyunwook Choi} \quad \textbf{Dahyun Chung} \quad \textbf{Hyunsung Kim} \quad \textbf{Siyoon Jin} \quad
\textbf{Jinhyeok Choi} \\ \textbf{Junyoung Seo} \quad \textbf{Seungryong Kim} \\[4pt]
KAIST AI \\[3pt]
\hypersetup{pdfborder={0 0 0}}\mdseries\href{https://cvlab-kaist.github.io/world-observer}{\textcolor{blue}{\texttt{https://cvlab-kaist.github.io/world-observer}}}
}

\newcommand{\paragrapht}[1]{\noindent\textbf{#1}}

\usepackage{booktabs}
\usepackage{graphicx}
\usepackage{wrapfig}

\iclrfinalcopy 
\begin{document}

\maketitle
\lhead{World Observer: Joint Actor-Observer Generation for Persistent World Modeling}  %

\begin{figure}[htbp]
    \centering
   \vspace{-20pt}
    \includegraphics[width=\linewidth]{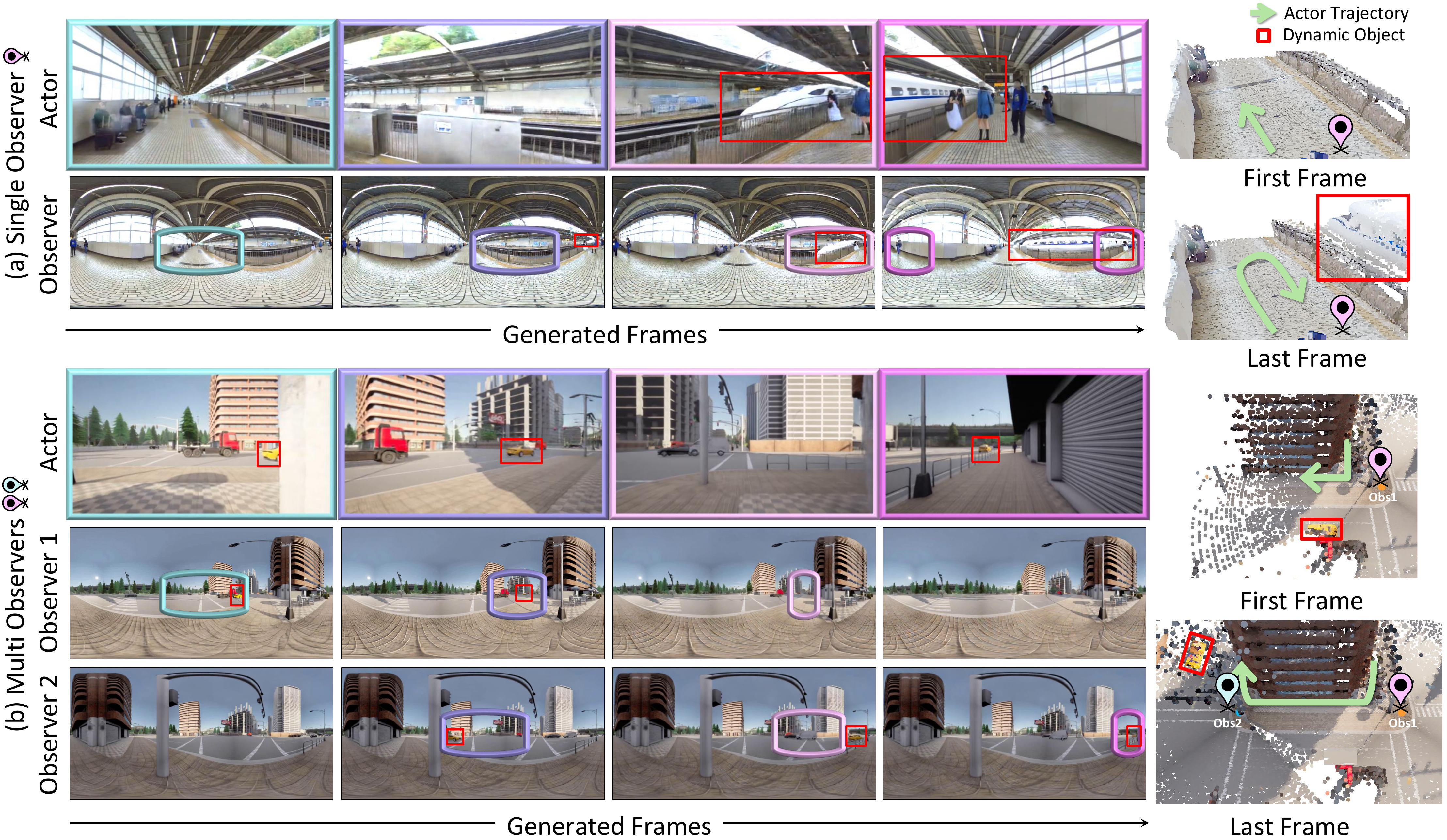}
    \vspace{-20pt}
\caption{
\textbf{Teaser.}
World Observer jointly generates panoramic observers with the actor. (a) Decoupled observer enables control over actor's unseen regions. (b) Multiple observers can be placed flexibly to track occluded regions.
}
\label{fig:teaser}
\end{figure}

\begin{abstract}

How can a world model continuously observe regions beyond the actor’s current view? Video world models simulate how an environment evolves from an agent's actions, yet remain actor-centric. Once an object leaves the actor's view, they lose direct evidence of its evolution, often failing to preserve its state and dynamics upon re-entry. To address this, we introduce \textbf{World Observer}, which decouples observing from acting by jointly generating a perspective \emph{actor} for the agent-centric view with one or more panoramic \emph{observers} that watch selected world regions. This allows objects that leave the actor's view to remain visually evolving in an observer, so their updated states are reflected when they re-enter. We ground the actor and observers by warping from a shared panoramic source for explicit geometric correspondence, and introduce an \emph{Observer Sink} of high-resolution perspective references to restore fine appearance upon re-entry. Since the observers are decoupled from the actor, they can be placed freely across the scene, extended to multiple locations for broader coverage, and driven by control signals to steer out-of-view evolution. To evaluate out-of-view evolution, we further introduce world-space metrics and a benchmark spanning real and synthetic scenes. World Observer substantially improves out-of-view dynamics while remaining competitive in visual fidelity, camera control, and 3D adherence.

\end{abstract}
    
\section{Introduction}
\label{sec:intro}
\begin{figure}[t!]       
    \centering    
    \includegraphics[width=\linewidth]{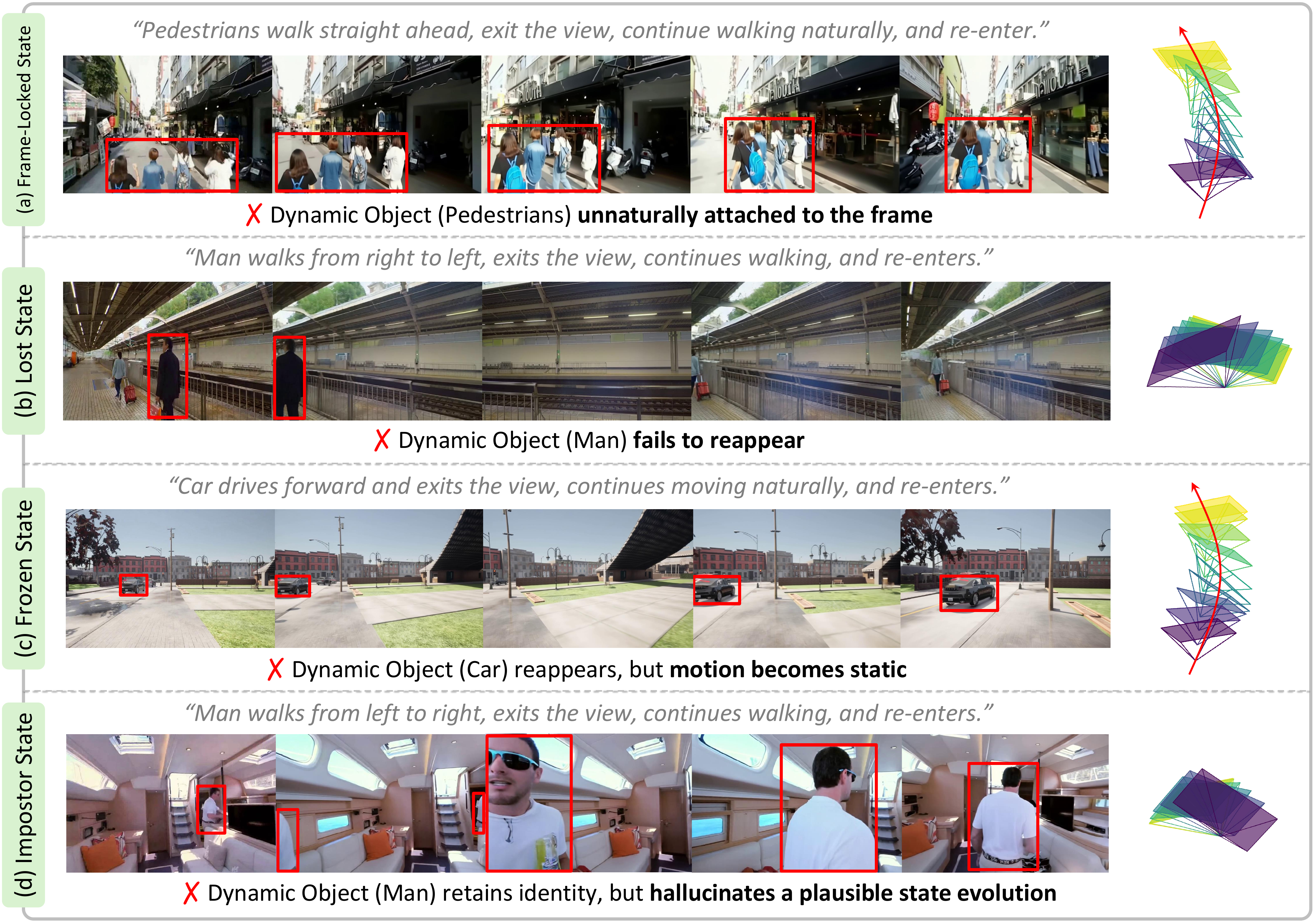}
    \vspace{-20pt}
    \caption{\textbf{Motivation.} When dynamic objects leave the actor's view, existing models exhibit four failures: (a) frame-locked, (b) lost, (c) frozen, and (d) impostor state. These failures show the difficulty of maintaining and evolving states without direct observations during out-of-view intervals.}
    \vspace{-15pt}
    \label{fig:motivation}   
\end{figure}

World models~\citep{nvidia2025worldsimulationvideofoundation, robbyantteam2026advancingopensourceworldmodels, hyworld2026hyworld20multimodalworld, dreamxteam2026dreamxworld10generalpurposeinteractive, wang2026matrixgame30realtimestreaming, dai2025fantasyworldgeometryconsistentworldmodeling, seo2026groundingworldsimulationmodels} simulate how environment evolves by generating future observations conditioned on an agent's actions. 
Existing models remain \textit{actor-centric}, updating the world primarily through what the actor currently observes, so whatever lies outside that view is left without direct visual evidence. A persistent world model should thus keep the world evolving regardless of where the actor looks, as regions outside the view can still affect the scene (Fig.~\ref{fig:teaser}(a)) and dynamic content keeps evolving even after it goes out of sight (Fig.~\ref{fig:teaser}(b)).
Such persistence matters for applications such as embodied navigation~\citep{li2025omninwmomniscientdrivingnavigation}, interactive simulation~\citep{robbyantteam2026advancingopensourceworldmodels, wang2026matrixgame30realtimestreaming, nvidia2025cosmosworldfoundationmodel, zhu2026astrageneralinteractiveworld, tang2026hunyuangamecraft2instructionfollowinginteractivegame, dreamxteam2026dreamxworld10generalpurposeinteractive}, and long-horizon planning~\citep{li2026magicworldlonghorizonstabilityinteractive}, where an agent must reason about a world beyond its current observation.

However, existing world models~\citep{nvidia2025worldsimulationvideofoundation, robbyantteam2026advancingopensourceworldmodels, hyworld2026hyworld20multimodalworld, dreamxteam2026dreamxworld10generalpurposeinteractive, wang2026matrixgame30realtimestreaming, dai2025fantasyworldgeometryconsistentworldmodeling, seo2026groundingworldsimulationmodels} often fail to maintain unobserved states. As illustrated in Fig.~\ref{fig:motivation}, we observe four characteristic failure modes that arise when an object moves out of the actor view. Fig.~\ref{fig:motivation}(a) shows a \textbf{frame-locked state}, where an object that should move out of view under camera motion instead remains unnaturally locked to the image frame. \citet{lu2026currentworldmodelslack} attributes this to models overemphasizing salient image-space subjects. Fig.~\ref{fig:motivation}(b) shows a \textbf{lost state}, where an object disappears after leaving the actor view and fails to reappear when the actor looks back. Fig.~\ref{fig:motivation}(c) shows a \textbf{frozen state}, where the object is retained but its dynamics stop evolving while unobserved, so it reappears near its last visible state. Fig.~\ref{fig:motivation}(d) shows an \textbf{impostor state}, where the object reappears with a plausible but incorrect state evolution that is inconsistent with what actually occurred while it was unobserved.

The core limitation is that world-state maintenance remains tied to the actor's observation, as which regions are continuously represented depends on where the actor looks. Once an object leaves the actor's observation, the model loses direct visual evidence of how its state evolves. Existing approaches compensate through internal memory~\citep{chen2026sightmindhybridmemory, wang2026worlddirector, wu2025videoworldmodelslongterm}, generative priors~\citep{xu2026teaching}, or explicit state extrapolation~\citep{duan2026liveworld}. However, these approaches infer the unseen evolution from previously observed information rather than directly observing what happens after the object leaves the view. As unobserved intervals grow longer or scene dynamics grow more complex, this inference becomes increasingly uncertain, causing state drift~\citep{lu2026currentworldmodelslack, ma2026sightmindevaluatingstate} and temporal inconsistency~\citep{chen2026sightmindhybridmemory, chen2026memobenchbenchmarkingworldmodeling, ma2026sightmindevaluatingstate}. The key challenge is therefore not merely to remember the last observed state, but to \emph{keep regions of interest observable regardless of where the actor looks}. This raises a natural question: \textit{How can a world model continuously observe regions beyond the actor’s current view?}

Our answer is to decouple \textit{observing} from \textit{acting}. We introduce \textbf{World Observer}, which jointly generates one or more \textit{observers} that watch chosen regions together with an \textit{actor} that renders the agent-centric view. To cover a broad surrounding region from a single observer location, we instantiate each observer as a $360^\circ$ panoramic video. Generated jointly, the actor and observers share a single world state, so a state seen by the actor continues in its observers and vice versa. Objects covered by an observer thus remain represented and evolve even outside the actor's view and reappear with updated states, mitigating the frame-locked, lost, frozen, and impostor failures.

A panoramic observer provides broad spatial coverage, but transferring its evolving world state to the perspective actor requires bridging geometric and appearance differences between the two representations. By warping the initial panorama into each viewpoint, we provide explicit geometric correspondences between overlapping regions of the actor and observer views. However, the observer's panoramic distortion limits the fine appearance detail available to the actor. We therefore introduce an \textit{Observer Sink}, a set of fixed high-resolution perspective references cropped from the initial panorama and accessed through shared attention. Together, the evolving observer and these appearance references help the actor render returning regions with updated states and fine detail.

Since observers are not tied to the actor, we can choose which regions to be observable independently of the actor's position and view. In Fig.~\ref{fig:teaser}(b), fixed multiple observers watch regions the actor cannot see, allowing occluded objects to remain tracked and re-enter consistently. Decoupling the observer from the actor also enables independent observer control, allowing us to guide how out-of-view regions evolve (Fig.~\ref{fig:teaser}(a)). Training such observers requires viewpoint-synchronized captures that are difficult to obtain in the real world, so we complement real panoramic videos with a synthetic CARLA~\citep{dosovitskiy2017carla} dataset that provides these observer configurations.

Finally, we evaluate whether objects continue to \emph{evolve} correctly while outside the actor view. Existing benchmarks~\citep{ma2026sightmindevaluatingstate, lu2026currentworldmodelslack, chen2026sightmindhybridmemory} largely focus on single, centered targets, while VLM-based metrics~\citep{lu2026currentworldmodelslack, ma2026sightmindevaluatingstate, chen2026memobenchbenchmarkingworldmodeling} mainly assess visual plausibility, so neither directly measures whether objects in moving, multi-object scenes evolve coherently while out of view. We therefore introduce world-space metrics that lift objects into 3D with a segmentation model~\citep{carion2026sam3segmentconcepts} and a metric-scale depth estimator~\citep{lin2025depth3recoveringvisual}. \textit{OOV-D}$_\text{gt}$ measures whether out-of-view motion agrees with ground-truth dynamics, \textit{OOV-D}$_\text{self}$ measures whether generated dynamics stay self-consistent through the interval, and \textit{OOV-F} reports valid leave-and-return cases. World Observer significantly improves out-of-view dynamics while remaining competitive on visual fidelity, camera control, and 3D adherence.

Our contributions are as follows.
\begin{itemize}
\item We introduce \textbf{World Observer}, which decouples acting from observing, generating a perspective actor with panoramic observers that keep the regions beyond the actor's view.
\item We decouple the observer from the actor, allowing us to freely place observers alongside the actor, at fixed viewpoints, or across scenes for broader coverage and out-of-view control.
\item We introduce \textit{OOV-D}$_\text{gt}$ and \textit{OOV-D}$_\text{self}$ for 3D out-of-view dynamic consistency, \textit{OOV-F} for valid leave-and-return cases, and a benchmark spanning real and synthetic scenes.
\end{itemize}

\section{Data Construction}
\begin{figure}[t!]       
    \centering    
    \includegraphics[width=\linewidth]{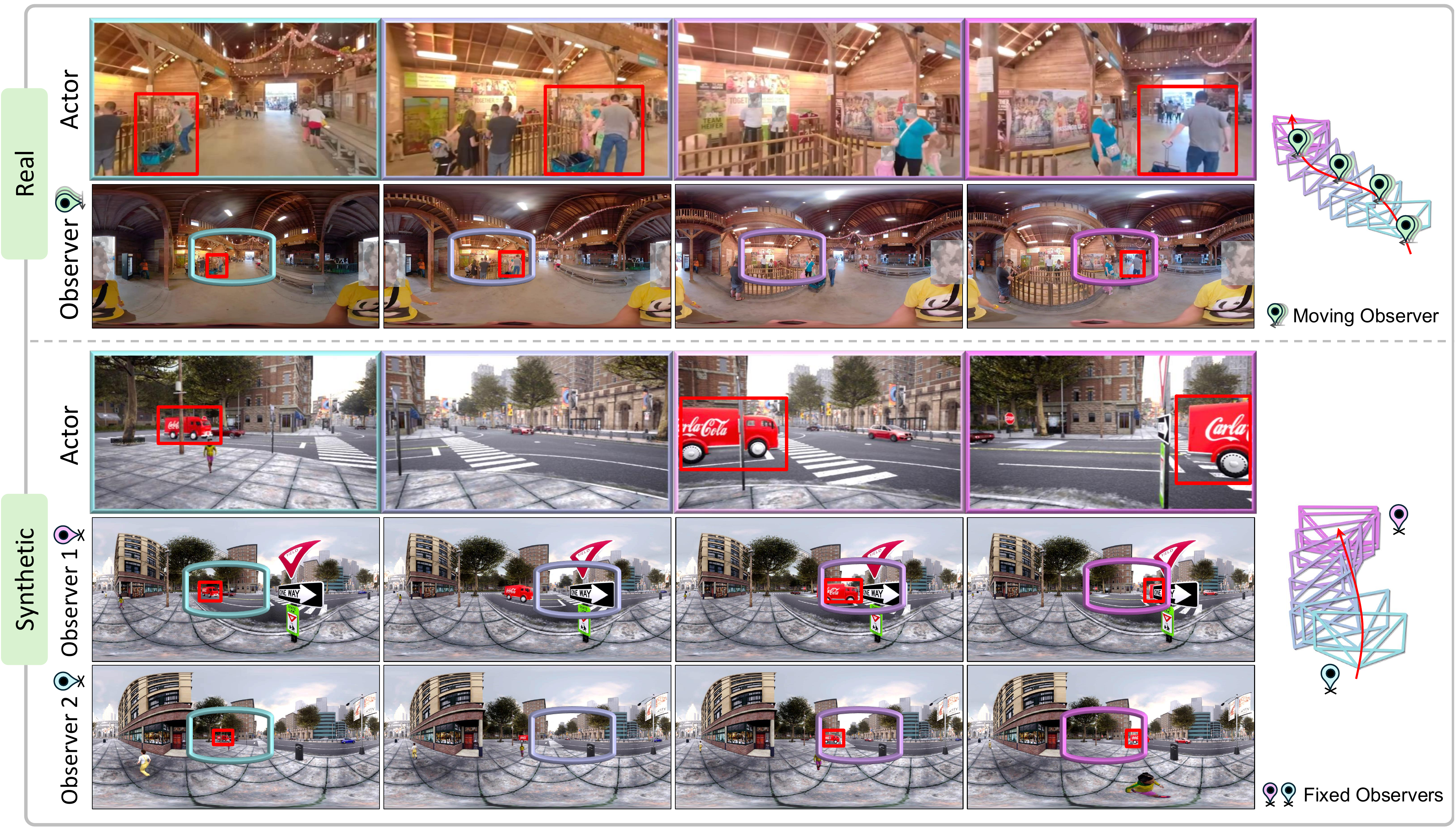}
    \vspace{-20pt}
    \caption{
\textbf{Dataset Overview.}
Real-world (top) and synthetic (bottom) examples with moving and two-fixed-observer configurations, respectively. Colored boxes show cross-view correspondences, while the rightmost diagrams show camera configurations and trajectories.}
    \label{fig:dataset_example}   
\end{figure}

\label{sec:data}
World Observer is trained on temporally synchronized actor and observer videos, built from two sources. Real panoramic video grounds generation in real-world appearance, while a synthetic simulator provides the decoupled and multi-observer configurations that real captures rarely contain.

\paragrapht{Real-world dataset.}
From an existing panorama corpus~\citep{luo2026framegenerating360panoramic, xia2025panowanliftingdiffusionvideo, chen2024360xpanopticmultimodalscene}, we curate high-resolution videos with stable motion and stabilize each into an upright, non-rotating equirectangular observer sequences. We then render synchronized perspective actor videos from the stabilized panorama under randomized trajectories and estimate per-frame depth with Depth Anything 3~\citep{lin2025depth3recoveringvisual} for the geometry-aware condition (Sec.~\ref{sec:compact}). Since both views come from the same panorama, real pairs are co-located, so we use synthetic data for observers decoupled from the actor and for multi-observer settings. Details are in Appendix~\ref{app:real_annotation}.

\paragrapht{Synthetic dataset.}
We render videos in CARLA~\citep{dosovitskiy2017carla} using the same stabilization and augmentation as the real data. The simulator lets us place observers independently of the actor, from one decoupled observer to several synchronized observers at different locations, providing complementary coverage (Fig.~\ref{fig:dataset_example}). Details are in Appendix~\ref{app:synthetic_annotation}.

\section{Method}
\label{sec:method}
\begin{figure}[t!]       
    \centering    
    \includegraphics[width=\linewidth]{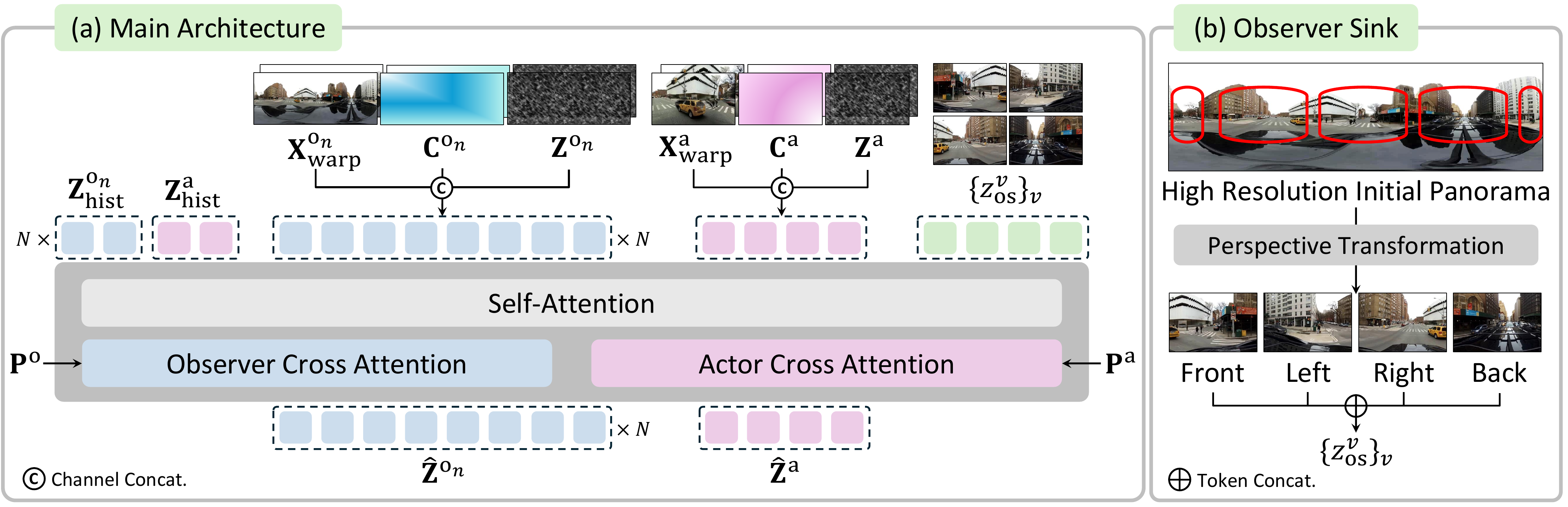}
    \vspace{-20pt}
\caption{
\textbf{Model Overview.}
(a) World Observer jointly generates actor and panoramic observer streams with a shared DiT.
(b) Observer Sink converts the high-resolution initial panorama into four perspective references for fine-grained appearance.
}
    \vspace{-15pt}
    \label{fig:main_architecture}   
\end{figure}

World Observer jointly generates video streams of two roles that share a single world. An \emph{actor} renders the agent's local view, while one or more panoramic \emph{observers} maintain broader world state and dynamics beyond it. Both streams are generated by a single pretrained video Diffusion Transformer (DiT)~\citep{nvidia2025worldsimulationvideofoundation} in the latent space of a 3D VAE, and each stream is conditioned on its own text prompt and camera trajectory. Long-horizon videos are generated autoregressively in chunks of $T$ frames, using tail latents of the previous chunk as history. Fig.~\ref{fig:main_architecture} shows the overview.

Given an initial actor view $x^{\mathrm a}_0$ and initial observer panoramas $\{x^{\mathrm o_n}_0\}_{n=1}^N$, World Observer generates an actor video $\mathbf X^{\mathrm a}{=}\{x^{\mathrm a}_t\}_{t=0}^{T-1}$ and $N$ observer videos $\{\mathbf X^{\mathrm o_n}\}_{n=1}^{N}$. For clarity, we present the single-observer case ($N{=}1$) and write $\mathbf X^{\mathrm o}$. For multi-observer ($N{>}1$), each observer has its own stream, as described in Sec.~\ref{sec:observer}. The actor stream takes a camera trajectory $\mathbf{C}^{\mathrm a}{=}\{c^{\mathrm a}_t\}_{t=0}^{T-1}$, a prompt $\mathbf{P}^{\mathrm a}$, and noisy latents $\mathbf{Z}^{\mathrm a}{=}\{z^{\mathrm a}_l\}_{l=0}^{L-1}$ to produce $\hat{\mathbf{Z}}^{\mathrm a}$, where $L$ is the number of latents per chunk. The observer stream is defined analogously with $\mathbf{C}^{\mathrm o}$, $\mathbf{P}^{\mathrm o}$, and $\mathbf{Z}^{\mathrm o}$, producing $\hat{\mathbf{Z}}^{\mathrm o}$. For each subsequent chunk, both streams additionally condition on history latents $\mathbf{Z}^{\mathrm a}_{\mathrm{hist}}$ and $\mathbf{Z}^{\mathrm o}_{\mathrm{hist}}$ from the tail of the previous chunk, where $H$ is the number of history latents.

\subsection{Joint Actor-Observer World Modeling}
\label{sec:joint}

While the actor cannot see beyond its view, the world keeps evolving. We therefore model the panorama as a time-varying \emph{observer} jointly generated with the actor, rather than a static reference. Objects leaving the actor's view keep evolving in the observer, and both streams share this state, so re-entering regions remain consistent with the observer instead of having its out-of-view evolution synthesized from scratch. The observer thus serves as an evolving world-state stream rather than static context.

\begin{wrapfigure}{r}{0.48\linewidth}
\centering
\includegraphics[width=\linewidth]{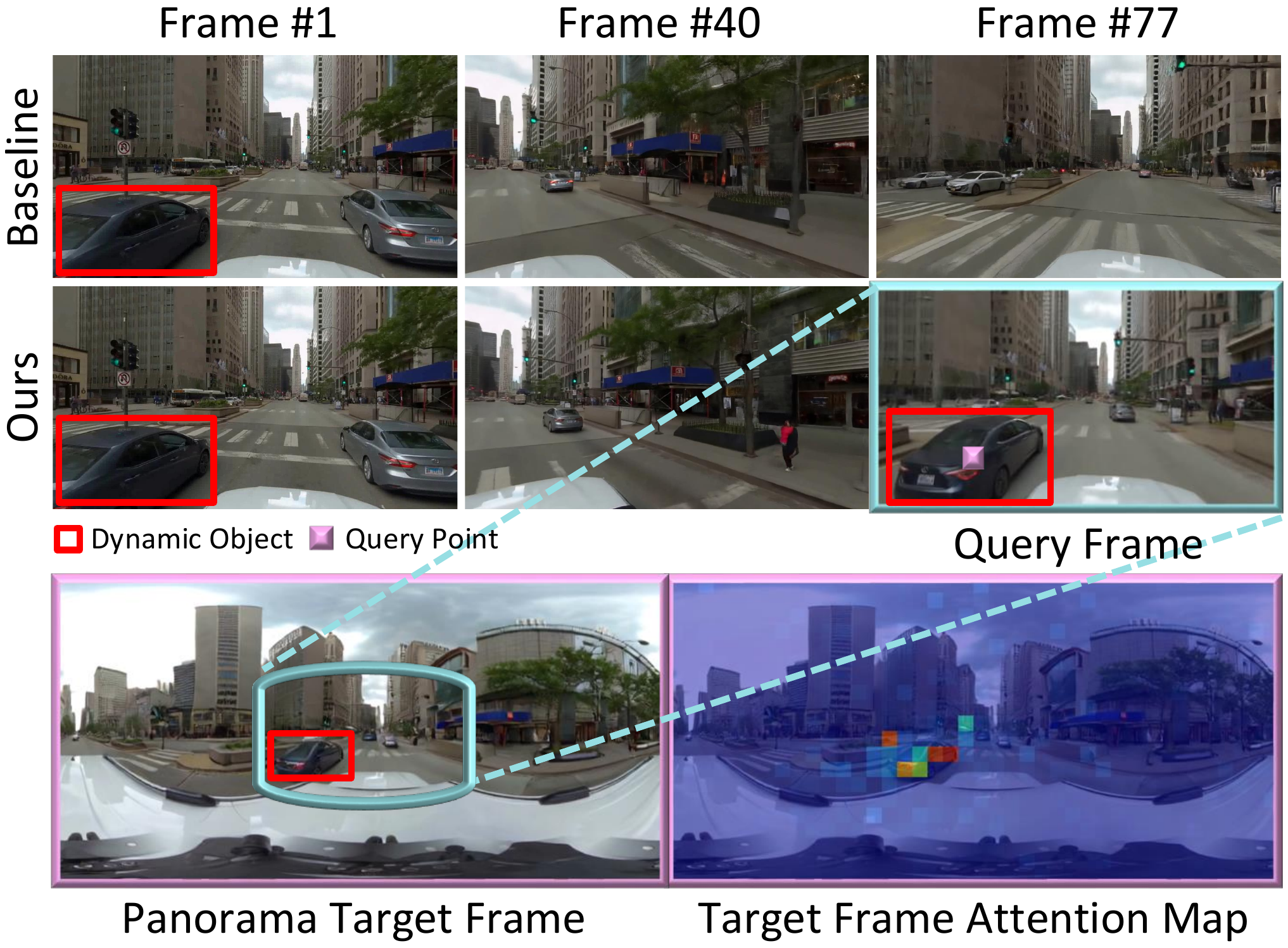}
\vspace{-20pt}
    \caption{\textbf{Visualization of self-attention.}
    A query on a re-entering object attends to the observer region holding its out-of-view state, while the object fails to re-enter after out-of-view interval in non-joint model (Baseline).}
    \vspace{-15pt}
    \label{fig:attention_viz}
\end{wrapfigure}

\paragrapht{View-time alignment.}
We concatenate actor and observer latents with their history latents from the previous chunk into a single sequence processed by shared DiT, adding learnable view embeddings to distinguish streams. Since both depict the same world at the same time, we assign matched actor and observer latents identical positions on the RoPE~\citep{su2023roformerenhancedtransformerrotary} temporal axis, allowing self-attention to exchange information across corresponding timesteps. As shown in Fig.~\ref{fig:attention_viz}, a query on a re-entering object can attend to the observer region at the same timestep, where its out-of-view state is maintained.

%sy
\paragrapht{Decoupled actor-observer prompting.}
The actor and the observer share one evolving world but describe different parts, so we condition them on separate prompts. The actor prompt $\mathbf{P}^{\mathrm a}$ specifies events within the actor's local view, while the observer prompt $\mathbf{P}^{\mathrm o}$ specifies events across the surrounding world, including regions the actor does not see. Since both streams are jointly generated and synchronized, this gives the observer two roles. It acts as \emph{memory}, since events leaving the actor view continue evolving in the observer, so a subject seen walking away keeps moving while unobserved. It also acts as \emph{control}, since the observer prompt can drive unseen events, such as an off-screen interaction, which the actor then observes when it looks there.

\paragrapht{Training objective.}
We train World Observer with flow matching~\citep{lipman2022flow}. For each stream $s\in\{\mathrm a,\mathrm o\}$, we sample Gaussian noise $\epsilon\sim\mathcal N(0,I)$ and a shared timestep $k\in[0,1]$, and construct the interpolated latent
\begin{equation}
    \tilde{\mathbf Z}^{s}_k=(1-k)\mathbf Z^s+k\epsilon.
\end{equation}
The objective is
\begin{equation}
    \mathcal{L}
    =
    \mathbb{E}\left[
    \sum_{s\in\{\mathrm a,\mathrm o\}}
    \left\|
    \mathbf v_\theta^s(\tilde{\mathbf Z}_{\mathrm{seq},k},\mathbf P^{\mathrm a},\mathbf P^{\mathrm o},k)
    -
    (\epsilon-\mathbf Z^s)
    \right\|_2^2
    \right],
\end{equation}
where $\tilde{\mathbf Z}_{\mathrm{seq},k}$ denotes the joint actor-observer sequence with interpolated target latents and clean history latents. The loss is applied only to the generated actor and observer latents.

\subsection{Panoramic Observer Modeling}
\label{sec:compact}
Since the panorama provides a shared view of the initial surroundings across all directions, we ground both streams in it through warping, fixing each stream's camera condition and aligning actor and observer geometry. Because the observer only has to maintain the world state rather than serving as the displayed output, we generate it at a flexible resolution. A panorama, however, inevitably carries geometric distortion, which loses fine appearance and weakens the correspondence between the observer and the actor. We therefore introduce an Observer Sink, a perspective reference from panorama that restores this appearance and correspondence when a region re-enters the actor view.

\paragrapht{Panoramic warping.}
Joint modeling shares information between the streams but does not fix the viewpoint each should render or how the actor and observer views correspond. We ground both in the initial panorama $x^{\mathrm o}_0$, whose $360^\circ$ coverage provide a shared view of the surrounding scene. Using its metric-scale depth $D_0$ from DA3~\citep{lin2025depth3recoveringvisual}, we warp this shared source into each stream's viewpoints along its own trajectory. For a target frame of stream $s\in\{\mathrm a,\mathrm o\}$ at pose $c^{s}_t$,
\begin{equation}
\label{eq:warping}
x^{s}_{\mathrm{warp},t} = \mathrm{Render}\bigl(\mathrm{Unproj}(x^{\mathrm o}_0,\, D^{\mathrm o}_0),\; c^{s}_{0\to t}\bigr),
\end{equation}
where $c^{s}_{0\to t}$ transforms the initial observer panorama to the target pose. Each warped video is encoded by the 3D VAE and channel-wise concatenated with the corresponding noisy latents and a binary validity mask marking covered pixels. Since both streams use the same panorama and depth, they are geometrically consistent by construction, giving shared attention a direct correspondence between the actor view and its observer region. We further channel-wise concatenate a Pl\"ucker raymap~\citep{sitzmann2022lightfieldnetworksneural} with each stream's latent features to encode its viewpoint.

\paragrapht{Decoupled resolution.}
Because the actor and observer play different roles, their resolutions need not match. The actor is the displayed output and must preserve visual detail, whereas the observer only maintains world state, so we generate it at a lower resolution $R^\mathrm{o}<R^\mathrm{a}$, where $R^\mathrm{a}$ is the actor resolution. Coarse spatial-temporal structure, object locations, motion, and visibility changes, is enough for the observer to track out-of-view dynamics, which keeps full-$360^\circ$ coverage tractable.

\paragrapht{Observer sink.}
As a distorted representation, the panorama maintains the state of the full surroundings but lacks the fine appearance the actor needs when it renders a re-entering region. To supply this, we introduce an \emph{Observer Sink}, a high-resolution perspective reference. We crop four perspective views from the high-resolution initial panorama, each $90^\circ$ apart in yaw, and encode each to a latent $z^{v}_{\mathrm{os}}$ for $v\in\{0,1,2,3\}$. The sink latents are kept fixed rather than denoised as target frames, and we append them to the actor-observer sequence so generated tokens can attend to them,
\begin{equation}
\label{eq:sink_seq}
\mathbf{Z}_\mathrm{seq} \leftarrow \bigl[\, \mathbf{Z}_\mathrm{seq};\; \{z^{v}_{\mathrm{os}}\}_{v} \,\bigr], \qquad
\mathbf{p}_\mathrm{seq} \leftarrow \bigl[\, \mathbf{p}_\mathrm{seq};\; \{\,H{+}L{+}\Delta_\mathrm{os}+v\,\delta_\mathrm{os}\,\}_{v} \,\bigr],
\end{equation}
where $\Delta_\mathrm{os}$ places the sink outside the generated-frame temporal position and $\delta_\mathrm{os}$ denotes the positional stride between sink views. The observer captures evolution over time, while the Observer Sink preserves appearance from the initial panorama, allowing the actor to render re-entering regions with updated states and fine detail.

\subsection{Flexible Observer Placement}
\label{sec:observer}

\paragrapht{Why flexible observers.}
An observer that simply follows the actor is constrained by the actor trajectory and cannot independently monitor a region of interest. However, important state updates may occur away from the actor, and such regions should remain observable regardless of where the actor moves. We therefore also consider a flexible single-observer setting, where the observer trajectory $\mathbf C^{\mathrm o}$ can be spatially separated from and move independently of the actor trajectory $\mathbf C^{\mathrm a}$, enabled by our synthetic data construction in Sec.~\ref{sec:data}. Beyond this, a single observer may still be insufficient when relevant regions are spatially separated or mutually occluded. We therefore further extend the formulation to $N>1$ observers, allowing multiple observer streams $\{\mathbf X^{\mathrm o_n}\}_{n=1}^{N}$ to maintain evolving state across different parts of the shared world.

\paragrapht{Multi-observer joint modeling.}
For $N>1$, we instantiate one panoramic stream for each observer $\mathbf X^{\mathrm o_n}$ and add an observer-specific learnable embedding to distinguish the streams. The actor stream and all $N$ observer streams are concatenated into a single sequence and jointly generated by the same DiT. For each actor trajectory frame, we select the initial observer panorama $x^{\mathrm o_n}_0$ whose corresponding observer pose is closest to the current actor position and use it as the geometric source for actor-view warping. This keeps the actor condition grounded in a nearby panorama for stable camera control while allowing the geometric reference to transition between observers as the actor moves through the scene. For each generation chunk, the Observer Sink is constructed from the closest observer selected from chunk's first frame, so its high-resolution appearance reference remains consistent with the geometric source. Meanwhile, all observer streams continue to evolve jointly, maintaining a coherent shared world across their different regions of coverage.

\section{Experiments}
\begin{figure}[t!]       
    \centering    
    \includegraphics[width=\linewidth]{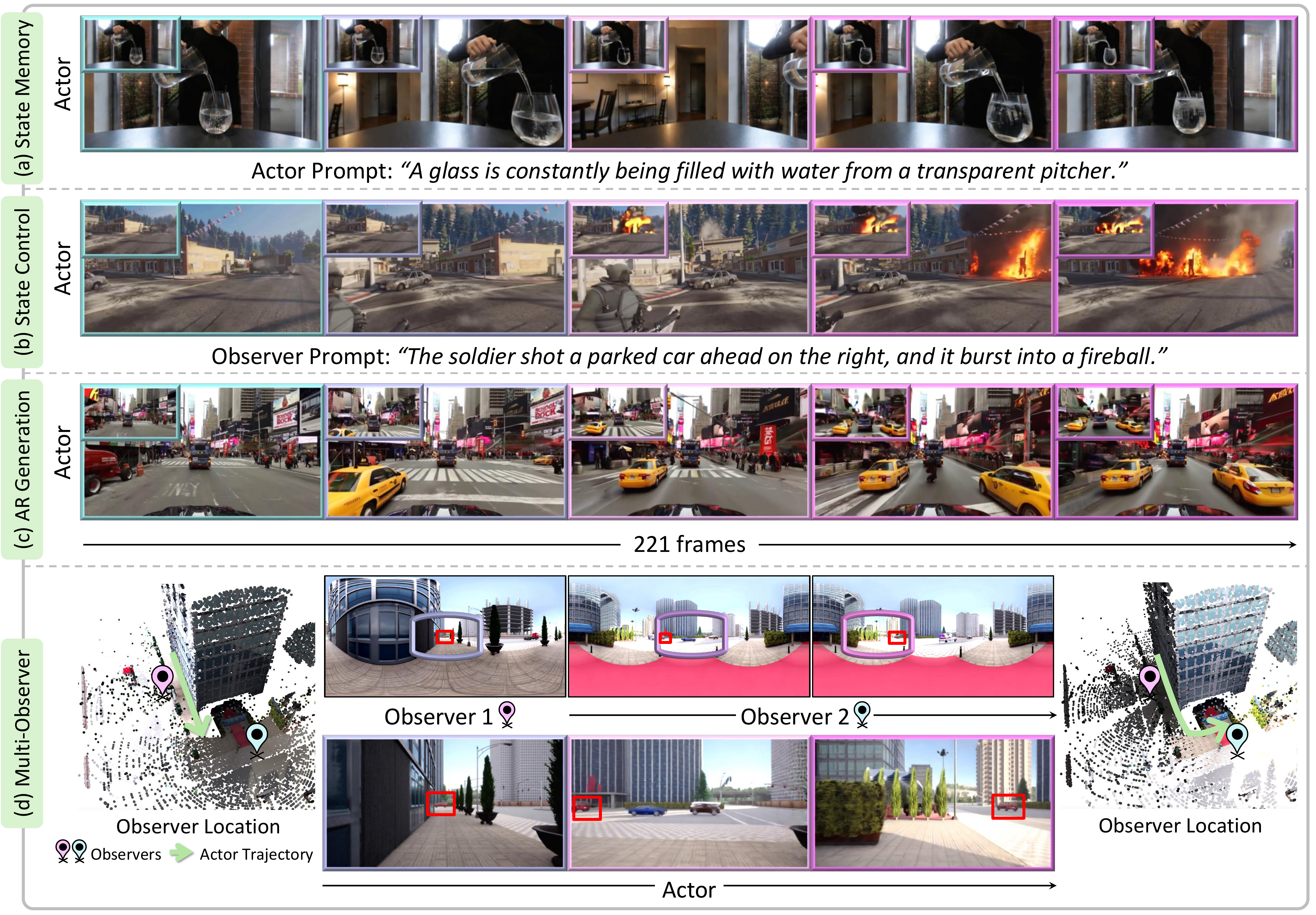}
    \vspace{-20pt}
\caption{
\textbf{Qualitative results.}
World Observer supports (a) state memory, (b) observer-prompted control, (c) long-horizon generation, and (d) multi-observer. Insets show the observer views.
}
    \vspace{-10pt}            
    \label{fig:qualitative_result}   
\end{figure}

\vspace{-5pt}
\subsection{Out-of-View Dynamics Metric}
\label{sec:metric}

Existing evaluations rely largely on semantic cues or VLM judgments, which struggle to quantify how multiple objects move while out-of-view. We instead measure out-of-view dynamics explicitly in world space. Using a segmentation model~\citep{carion2026sam3segmentconcepts} and a metric-scale depth estimator~\citep{lin2025depth3recoveringvisual}, we lift each object into 3D and track its position over time, turning unobserved motion into real-world displacement with comparable direction and magnitude across methods.

We measure two aspects of displacement for objects that validly exit and re-enter the actor's view. \textbf{OOV-F} reports how often objects follow the ground-truth exit-and-re-entry pattern, capturing the \textit{frame-locked} and \textit{lost} failures with no valid displacement. For objects that exit and return, \textbf{OOV-D} measures how well out-of-view motion matches a reference. We use two references. \textbf{OOV-D}$_\text{gt}$ compares motion with the ground-truth dynamics specified by the prompt, while \textbf{OOV-D}$_\text{self}$ checks whether the object continues pre-exit motion. Accordingly, OOV-D$_\text{gt}$ is more sensitive to frozen motion, whereas OOV-D$_\text{self}$ is more sensitive to impostor motion. Details are in Appendix~\ref{app:oov_metrics}.

\subsection{Implementation Details}
\paragraph{Model and training setup.}
World Observer fine-tunes Cosmos-Predict2.5~\citep{nvidia2025worldsimulationvideofoundation} with AdamW~\citep{loshchilov2019decoupledweightdecayregularization} at a learning rate of $4.8\mathrm{e}{-5}$ on 8 NVIDIA H100 GPUs. We train on 138K clips sampled with diverse trajectories from 23K real videos~\citep{luo2026framegenerating360panoramic, xia2025panowanliftingdiffusionvideo, chen2024360xpanopticmultimodalscene} and with 72K clips from 12K synthetic videos~\citep{dosovitskiy2017carla}, with $T{=}77$ frames. The single-observer model trains for 12K iterations with batch size 16. The multi-observer model starts from this checkpoint and is trained for additional 6K iterations with total batch size 8 on synthetic data using $N{=}2$ observers. We render actor at $R^\mathrm{a}{=}1280\times704$ and observer at $R^\mathrm{o}{=}640\times320$, with Observer Sink offset $\Delta_\mathrm{os}{=}50$. Further details are in Appendix~\ref{app:training}.

\paragraph{Evaluation benchmarks.}
We build held-out benchmarks on real panoramic video and synthetic CARLA video, unseen during training, with $100$ sequences each ($200$ in total) of $T{=}77$ frames. We evaluate within a single chunk, where the full sequence fits in context and no memory retrieval is needed. This isolates whether the model genuinely maintains out-of-view dynamics, since under an unlimited-memory single chunk a failure cannot be attributed to limited context or missing retrieval. Each sequence uses a back-and-forth camera rotation with random angle and speed and contains one or more objects undergoing natural out-of-view motion, also provided as a text prompt.

\paragraph{Baselines.}
We compare against general video world models~\citep{robbyantteam2026advancingopensourceworldmodels, dreamxteam2026dreamxworld10generalpurposeinteractive, wang2026matrixgame30realtimestreaming, dai2025fantasyworldgeometryconsistentworldmodeling, hyworld2025}, panorama world models~\citep{liu2026omniroamworldwanderinglonghorizon, li2026panoworldrealworldpanoramicgeneration}, and the out-of-view method~\citep{chen2026sightmindhybridmemory}. Since these models do not all accept our inputs, we provide each baseline with its supported conditions. For panorama world models, whose output is a full panorama rather than a perspective view, we crop a perspective view from the output using the ground-truth camera rotation, making their output directly comparable to our perspective actor. Additional adaptation details are in Appendix~\ref{app:baselines}.

\paragraph{Metrics.}
Visual and temporal quality use FID~\citep{heusel2018ganstrainedtimescaleupdate} and FVD~\citep{unterthiner2019accurategenerativemodelsvideo}, and VBench~\citep{huang2023vbenchcomprehensivebenchmarksuite} Image Quality. \textbf{Camera-following accuracy} uses Rotation Error (RotErr) and Translation Error (TransErr). \textbf{3D adherence} uses masked PSNR and LPIPS~\citep{zhang2018unreasonableeffectivenessdeepfeatures} on static regions segmented with SAM3~\citep{carion2026sam3segmentconcepts}, since generated dynamics need not match the ground truth. \textbf{Out-of-view dynamics} is measured with $\text{OOV-F}$, $\text{OOV-D}_\text{gt}$, and $\text{OOV-D}_\text{self}$, defined in Sec.~\ref{sec:metric}, where OOV-F reports how often objects validly exit and re-enter and the two OOV-D scores measure how their out-of-view motion evolves.

\subsection{Results}
\label{sec:results}
\begin{figure}[t!]       
    \centering    
    \includegraphics[width=\linewidth]{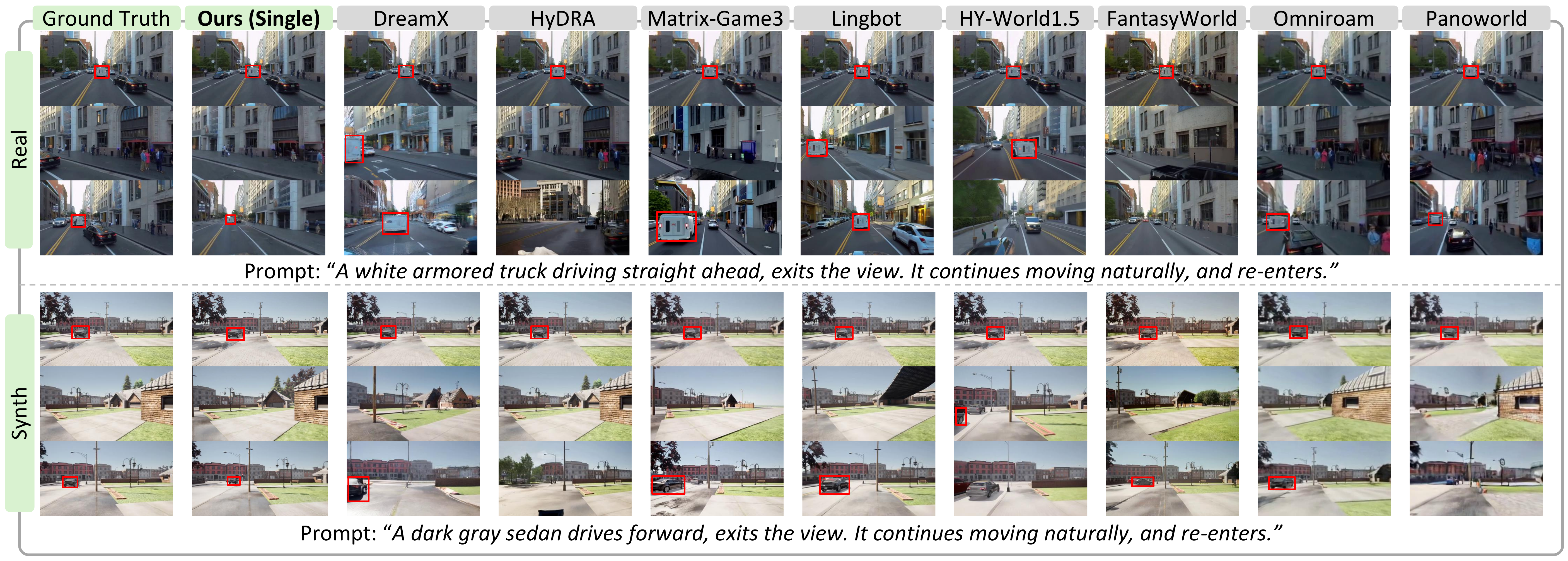}
    \vspace{-20pt}
    \caption{\textbf{Qualitative comparisons with other models.} Red boxes mark target objects before exit and after re-entry. While existing models often lose, freeze, or inconsistently reconstruct them, World Observer better preserves object identity and state evolution across out-of-view intervals.}  
    \vspace{-15pt}            
    \label{fig:qualitative_comparison}   
\end{figure}

\begin{table}[t]
    \centering
    \caption{
        \textbf{Quantitative comparison with existing methods.}
        Results are reported as
        Real-OOV-Bench / Synthetic-OOV-Bench.
        RotErr is (-) for panorama outputs cropped with the ground-truth rotation.
        World Observer (Multi) is evaluated only on Synthetic-OOV-Bench, as multi-observer conditions are scarce in the real world.
    }
    \label{tab:main_quan}
    \setlength{\tabcolsep}{3pt}
    \resizebox{\columnwidth}{!}{%
        \begin{tabular}{l|c|ccc|cc|cc|ccc}
            \toprule
            &
            & \multicolumn{3}{c|}{Visual \& Temporal Fidelity}
            & \multicolumn{2}{c|}{Camera Control}
            & \multicolumn{2}{c|}{3D Adherence}
            & \multicolumn{3}{c}{Out-of-View Dynamics} \\
            \cmidrule(lr){3-5}
            \cmidrule(lr){6-7}
            \cmidrule(lr){8-9}
            \cmidrule(lr){10-12}
            Method
            & \#Param
            & FID$\downarrow$
            & FVD$\downarrow$
            & Img.Q.$\uparrow$
            & RotErr$\downarrow$
            & TransErr$\downarrow$
            & mPSNR$\uparrow$
            & mLPIPS$\downarrow$
            & OOV-F$\uparrow$
            & OOV-D$_\text{gt}\uparrow$
            & OOV-D$_\text{self}\uparrow$ \\
            \midrule
            \multicolumn{12}{c}{\textit{Panorama World Models}} \\
            \midrule
            OmniRoam~\citep{liu2026omniroamworldwanderinglonghorizon}
            & 1.3B
            & 42.13 / 31.56
            & \underline{404.7} / 418.9
            & 0.387 / 0.390
            & - / -
            & 0.616 / 0.072
            & \underline{18.523} / 15.934
            & \underline{0.317} / 0.388
            & \underline{0.563} / \textbf{0.875}
            & 0.055 / 0.107
            & 0.063 / 0.211 \\
            PanoWorld~\citep{li2026panoworldrealworldpanoramicgeneration}
            & 5B
            & 62.48 / 55.73
            & 677.4 / 557.2
            & 0.397 / 0.377
            & - / -
            & 0.659 / 0.090
            & 17.055 / 15.426
            & 0.383 / 0.418
            & 0.476 / 0.289
            & 0.177 / 0.292
            & 0.142 / 0.216 \\
            \midrule
            \multicolumn{12}{c}{\textit{Video World Models}} \\
            \midrule
            HyDRA~\citep{chen2026sightmindhybridmemory}
            & 1.3B
            & 50.99 / 67.94
            & 615.9 / 944.4
            & 0.637 / 0.656
            & 0.140 / 0.164
            & 0.561 / 0.510
            & 14.859 / 13.455
            & 0.407 / 0.490
            & 0.416 / 0.258
            & 0.274 / 0.266
            & 0.292 / 0.253 \\
            HY-World 1.5~\citep{hyworld2025}
            & 8B
            & 51.61 / 42.63
            & 518.7 / 509.1
            & 0.628 / 0.640
            & 0.479 / 0.419
            & 0.511 / 0.122
            & 12.690 / 13.012
            & 0.517 / 0.531
            & 0.128 / 0.133
            & 0.358 / 0.351
            & 0.311 / 0.311 \\
            Dream-X~\citep{dreamxteam2026dreamxworld10generalpurposeinteractive}
            & 5B
            & 42.91 / 38.95
            & 470.5 / 635.6
            & 0.646 / 0.651
            & 0.138 / 0.110
            & \underline{0.478} / 0.066
            & 14.054 / 13.572
            & 0.473 / 0.522
            & 0.484 / 0.352
            & 0.280 / 0.308
            & 0.213 / 0.310 \\
            FantasyWorld~\citep{dai2025fantasyworldgeometryconsistentworldmodeling}
            & 14B
            & 45.55 / 35.72
            & 511.9 / 385.3
            & 0.641 / \underline{0.679}
            & 0.197 / 0.123
            & 0.616 / 0.105
            & 13.241 / 13.848
            & 0.472 / 0.481
            & 0.288 / 0.633
            & \underline{0.382} / 0.338
            & \underline{0.379} / 0.390 \\
            LingBot-World~\citep{robbyantteam2026advancingopensourceworldmodels}
            & 28B
            & \underline{39.87} / 28.75
            & 420.0 / 333.3
            & \textbf{0.668} / 0.675
            & 0.249 / 0.160
            & 0.548 / 0.098
            & 13.329 / 14.094
            & 0.493 / 0.493
            & 0.333 / 0.570
            & 0.337 / 0.340
            & 0.343 / \underline{0.460} \\
            Matrix-Game-3~\citep{wang2026matrixgame30realtimestreaming}
            & 5B
            & 49.41 / 33.00
            & 541.1 / 393.7
            & \underline{0.659} / \textbf{0.689}
            & \textbf{0.089} / 0.053
            & 0.589 / 0.148
            & 11.890 / 12.354
            & 0.507 / 0.517
            & \textbf{0.568} / 0.508
            & 0.066 / 0.125
            & 0.089 / 0.118 \\
            \midrule
            \textbf{World Observer (Single)}
            & 2B
            & \textbf{29.30} / \underline{19.65}
            & \textbf{291.7} / \underline{196.2}
            & 0.647 / 0.641
            & \underline{0.125} / \underline{0.022}
            & \textbf{0.372} / \underline{0.029}
            & \textbf{18.574} / \underline{18.893}
            & \textbf{0.278} / \underline{0.276}
            & 0.492 / 0.580
            & \textbf{0.426} / \textbf{0.531}
            & \textbf{0.408} / \textbf{0.526} \\
            \textbf{World Observer (Multi)}
            & 2B
            & - / \textbf{16.06}
            & - / \textbf{147.0}
            & - / 0.650
            & - / \textbf{0.014}
            & - / \textbf{0.022}
            & - / \textbf{20.357}
            & - / \textbf{0.240}
            & - / \underline{0.722}
            & - / \underline{0.528}
            & - / 0.457 \\
            \bottomrule
        \end{tabular}%
    }
    \vspace{-15pt}
\end{table}

\paragrapht{Generation results.}
Fig.~\ref{fig:qualitative_result} shows the capabilities of World Observer. (a) \textbf{State memory}: region that leaves the actor's view keeps evolving in the observer and re-enters with a consistent, updated state rather than being lost, frozen, or hallucinated. (b) \textbf{State controllability}: the observer prompt drives objects outside the actor's field of view, and triggered events continue unfolding until re-entry. (c) \textbf{Autoregressive generation}: World Observer rolls out chunk by chunk over long-horizon while keeping the actor and observer consistent. (d) \textbf{Multi-Observer generation}: observers placed at specific locations cover regions occluded from the actor, so occluded objects remain tracked and re-enter consistently.

\paragrapht{Comparison with other models.}
As shown in Tab.~\ref{tab:main_quan} and Fig.~\ref{fig:qualitative_comparison}, World Observer performs favorably in visual and temporal fidelity, camera-following accuracy, and 3D adherence, while achieving the strongest out-of-view dynamics. The metrics also reveal baseline failures. Low OOV-F means objects are frame-locked or lost and never validly exit and re-enter. High OOV-F with low OOV-D$_\text{gt}$ and OOV-D$_\text{self}$ means objects re-enter but match neither reference, being frozen or moving incoherently. High OOV-D$_\text{gt}$ with lower OOV-D$_\text{self}$ means impostor dynamics, where motion follows the prompt but departs from the object's own trajectory. World Observer keeps both OOV-F and OOV-D high, so its re-entering objects keep a consistent, continuing state instead of freezing or turning into impostors. Adding observers further raises OOV-F by covering more regions of interest, while both OOV-D metrics remain competitive.

\subsection{Ablation Study}
\label{sec:analysis}
Representative qualitative ablations of the conditioning strategies are shown in Appendix~\ref{app:qual_ablation}.

\paragrapht{Joint generation.}
As shown in Tab.~\ref{tab:ablation}, the Backbone, which uses only raymap conditioning, shows unstable camera control with the lowest OOV scores overall. Removing the observer generates everything from the actor alone, and its relatively high OOV-F alongside low OOV-D$_\text{gt}$ and OOV-D$_\text{self}$ indicates that the actor tends to bring previously unseen objects back into frame but with an inconsistent state, an impostor tendency. Generating without the actor instead handles out-of-view content directly in the panorama and therefore reaches strong OOV scores, but the panorama is inherently distorted, so its camera control and visual fidelity fall short. Generating the actor and observer jointly combines these strengths, achieving the highest OOV-D with a stable OOV-F while remaining strong in camera control, 3D adherence, and visual fidelity.

\paragraph{Observer sink.}
Without the Observer Sink, the actor and observer have no shared high-resolution perspective reference of the surroundings, so less appearance information is available to ground the generated content. This is reflected in the lower OOV-D$_\text{self}$, and the weaker exchange of surrounding information also degrades 3D adherence, FID, and FVD.

\begin{table}[t] % WorldObserver: 2B
    \centering
    \caption{
        \textbf{Ablation study on Synthetic-OOV-Bench.}
        We evaluate visual and temporal fidelity, camera-following accuracy, 3D adherence, and out-of-view dynamics. For fairness, all variants share the full model's conditioning. RotErr is (-) for w/o Actor generation, whose panoramas are cropped with the ground-truth rotation.
    }
    \label{tab:ablation}
    \setlength{\tabcolsep}{4pt}
    \resizebox{\columnwidth}{!}{%
        \begin{tabular}{l|ccc|cc|cc|ccc}
            \toprule
            & \multicolumn{3}{c|}{Visual \& Temporal Fidelity}
            & \multicolumn{2}{c|}{Camera Control}
            & \multicolumn{2}{c|}{3D Adherence}
            & \multicolumn{3}{c}{Out-of-View Dynamics} \\
            \cmidrule(lr){2-4}
            \cmidrule(lr){5-6}
            \cmidrule(lr){7-8}
            \cmidrule(lr){9-11}
            Variant
            & FID$\downarrow$
            & FVD$\downarrow$
            & Img.Q.$\uparrow$
            & RotErr$\downarrow$
            & TransErr$\downarrow$
            & mPSNR$\uparrow$
            & mLPIPS$\downarrow$
            & OOV-F$\uparrow$
            & OOV-D$_\text{gt}\uparrow$
            & OOV-D$_\text{self}\uparrow$ \\
            \midrule
            Backbone (raymap only)
            & 36.59
            & 450.5
            & \textbf{0.674}
            & 0.595
            & 0.134
            & 12.463
            & 0.543
            & 0.070
            & 0.332
            & 0.414 \\
            w/o Observer generation
            & 24.82
            & 243.1
            & \underline{0.659}
            & 0.057
            & 0.073
            & 16.997
            & 0.337
            & \underline{0.674}
            & 0.387
            & 0.430 \\
            w/o Actor generation
            & 21.38
            & 259.9
            & 0.433
            & -
            & 0.048
            & 18.591
            & 0.315
            & 0.554
            & 0.471
            & \underline{0.499} \\
            w/o Observer Sink
            & 20.33
            & 214.0
            & 0.646
            & 0.026
            & 0.031
            & 18.221
            & 0.303
            & 0.545
            & 0.506
            & 0.470 \\
            \midrule
            \textbf{World Observer (Single)}
            & \underline{19.65}
            & \underline{196.2}
            & 0.641
            & \underline{0.022}
            & \underline{0.029}
            & \underline{18.893}
            & \underline{0.276}
            & 0.580
            & \textbf{0.531}
            & \textbf{0.526} \\
            \textbf{World Observer (Multi)}
            & \textbf{16.06}
            & \textbf{147.0}
            & 0.650
            & \textbf{0.014}
            & \textbf{0.022}
            & \textbf{20.357}
            & \textbf{0.240}
            & \textbf{0.722}
            & \underline{0.528}
            & 0.457 \\
            \bottomrule
        \end{tabular}%
    }
    \vspace{-15pt}
\end{table}

\section{Conclusion}
We presented \textbf{World Observer}, which decouples observing from acting by jointly generating a perspective actor with panoramic observers that maintain world state beyond the actor's view. This formulation keeps out-of-view regions explicitly represented as the world evolves, allowing their updated states to be reflected when they later re-enter the actor view. Panorama grounding and the Observer Sink preserve geometric consistency and fine appearance across actor and observer views, while decoupled observers enable flexible placement and control of out-of-view regions. We further introduced world-space metrics and benchmarks on real and synthetic scenes to directly evaluate this capability. Across these evaluations, World Observer better preserves out-of-view dynamics while remaining competitive in visual fidelity, camera control, and 3D adherence.

\bibliography{iclr2027_conference}
\bibliographystyle{iclr2027_conference}

\appendix
\newpage
\section*{\Large Appendix}
{
    \renewcommand{\baselinestretch}{1.8}\selectfont
    
    \noindent
    \textbf{A\quad Related Work}
    
    \hspace*{2em}A.1\quad Video World Models
    \vspace{-5pt}
    
    \hspace*{2em}A.2\quad Panorama World Models
    \vspace{-5pt}
    
    \hspace*{2em}A.3\quad Out-of-State and Dynamics 
    
    \noindent\textbf{B\quad Implementation Details}
    
    \hspace*{2em}B.1\quad Model and Training Details
    \vspace{-5pt}
    
    \hspace*{2em}B.2\quad Real Data Processing
    \vspace{-5pt}
    
    \hspace*{2em}B.3\quad Synthetic Data Processing
    
    \noindent\textbf{C\quad Evaluation Details}
    
    \hspace*{2em}C.1\quad Benchmark Construction
    \vspace{-5pt}
    
    \hspace*{2em}C.2\quad Evaluation Metrics
    \vspace{-5pt}
    
    \hspace*{2em}C.3\quad Out-of-View Dynamics Metrics
    \vspace{-5pt}
    
    \hspace*{2em}C.4\quad Comparison with existing Out-of-View Metrics
    \vspace{-5pt}
    
    \hspace*{2em}C.5\quad Baseline Adaptation
    
    \noindent\textbf{D\quad Additional Results and Analyses}
    
    \hspace*{2em}D.1\quad Additional Qualitative Results
    \vspace{-5pt}
    
    \hspace*{2em}D.2\quad Additional Qualitative Ablation Results
    
    \noindent\textbf{E\quad Discussions}
    
    \hspace*{2em}E.1\quad Limitations
    \vspace{-5pt}

    \hspace*{2em}E.2\quad Societal Impact
    
    \vspace{10pt}
}

\vfill

\clearpage

\section{Related Work}
\subsection{Video World Models}
Building on recent advances in video generation~\citep{wan2025wanopenadvancedlargescale, zheng2024opensorademocratizingefficientvideo, kong2025hunyuanvideosystematicframeworklarge, yang2025cogvideoxtexttovideodiffusionmodels, yin2025slowbidirectionalfastautoregressive, huang2025selfforcingbridgingtraintest}, video world models adopt generative models to predict future observations from an actor-centric view. Some video world models~\citep{nvidia2025worldsimulationvideofoundation, robbyantteam2026advancingopensourceworldmodels} rely primarily on the generative prior learned from large-scale pretraining. To maintain a coherent world beyond the actor currently observes, recent approaches additionally retrieve relevant past observations~\citep{hyworld2026hyworld20multimodalworld, yu2025contextmemorysceneconsistentinteractive, dreamxteam2026dreamxworld10generalpurposeinteractive, wang2026matrixgame30realtimestreaming, xiao2026worldmemlongtermconsistentworld, yi2026worldkvefficientworldmemory} or maintain stored states in explicit spatial and geometric representations~\citep{wu2025videoworldmodelslongterm, li2025vmemconsistentinteractivevideo, yu2026mosaicmemhybridspatialmemory, li2026magicworldlonghorizonstabilityinteractive}. However, generative priors, retrieved observations, and stored states still lack direct evidence of how dynamic content evolves once it leaves the actor view, leading to the \textit{frame-locked}, \textit{lost}, \textit{frozen}, and \textit{impostor} failures shown in Fig.~\ref{fig:motivation}. In contrast, we jointly generate a panoramic \textit{observer} that continues to represent the evolving world beyond the actor's view, so that out-of-view dynamic content is explicitly generated.
\subsection{Panorama World Models}
To overcome the limited FoV of perspective models, prior work has explored panoramas either as intermediate representations for constructing coherent 3D worlds~\citep{wang2025evoworldevolvingpanoramicworld, schnepf2026sphericaldreamergeneratingnavigableimmersive, hyworld2026hyworld20multimodalworld} or as direct world representations through panoramic video generation~\citep{yin2025panoworldxgeneratingexplorablepanoramic, li2026panoworldrealworldpanoramicgeneration, chen2026pantheon360tamingdigitaltwin, liu2026omniroamworldwanderinglonghorizon}. Producing panoramas as the final output introduces additional challenges: pretrained backbones are typically optimized for perspective imagery, which has motivated tangent-plane or cubemap-based formulations~\citep{kalischek2025cubediffrepurposingdiffusionbasedimage, çapuk2025tandittangentplanediffusiontransformer}, while representing the full sphere under a fixed token budget can reduce the effective resolution available to each viewing direction~\citep{li2026cubecomposerspatiotemporalautoregressive4k, liu2025dynamicscalerseamlessscalablevideo}. In contrast, we decouple observing from acting within an actor-observer framework, allowing one or more observers to observe selected regions independently of the actor’s position and view, while the perspective actor is jointly generated as the primary output. Because the observer maintains the surrounding world rather than serving as the actor view, it need not carry the actor's fidelity, which sidesteps these resolution and prior limitations while the perspective actor stays high-quality.
\subsection{Out-of-View State and Dynamics}
Recent benchmarks examine whether world models maintain evolving states when parts of the world become temporarily unobserved. STEVO-Bench~\citep{ma2026sightmindevaluatingstate} tests state and dynamics consistency under controlled observation interruptions. WRBench~\citep{lu2026currentworldmodelslack} evaluates state consistency when the camera moves away from a scene and later returns. MemoBench~\citep{chen2026memobenchbenchmarkingworldmodeling} evaluates whether out-of-view objects re-enter with appropriately evolved states across diverse dynamic scenes, including physical state changes. These benchmarks, however, leave the judgment to a VLM, which rates whether a reappearance looks natural. In the dynamic, multi-object scenes world models produce, naturalness depends on where an object actually moved while unobserved, which such judgments cannot verify.
A few approaches address interrupted observation directly. HyDRA~\citep{chen2026sightmindhybridmemory} and ReMind~\citep{xu2026teaching} use dynamic memory and generative priors to continue unseen dynamics, while LiveWorld~\citep{duan2026liveworld} and WorldDirector~\citep{wang2026worlddirector} explicitly advance hidden states through simulation or motion planning. Yet generative approaches may drift toward plausible but incorrect unseen states, memory-based methods remain limited by previously observed evidence, and explicit state advancement requires structured entity-level modeling. We instead use a panoramic video as an evolving global observer that continuously captures out-of-view dynamics.

\section{Implementation Details}
\label{app:implementation}

\subsection{Model and Training Details}
\label{app:training}

\subsubsection{Additional training details}
World Observer is trained with teacher forcing (TF) throughout. During TF training, we
inject small Gaussian noise (scale $0.1$) to the conditioning history frames with $50\%$
probability, reducing the gap between clean training inputs and self-generated inference
history. Each chunk conditions on $H{=}2$ history latents, corresponding to $5$ frames of
the preceding chunk. The two training data sources are mixed via ratio-based interleaved
sampling: real panoramic videos~\citep{luo2026framegenerating360panoramic,
xia2025panowanliftingdiffusionvideo, chen2024360xpanopticmultimodalscene} ($60\%$) and synthetic
video from CARLA~\citep{dosovitskiy2017carla} ($40\%$). The multi-observer stage is trained on synthetic data only, as synchronized multi-observer recordings are difficult to obtain in the real world. All camera extrinsics are converted to a unified Right-Down-Forward (RDF) coordinate system and expressed relative to the first frame of each chunk. 

During training, warping source is not fixed to the first frame of the training clip. We sample it from the full source video surrounding the clip, taking the nearest past or future frame beyond a minimum spatial distance from the target, so the condition is rarely a clean, well-aligned view of the target. This matches the autoregressive setting, where later chunks are warped from frames that are increasingly distant from the current viewpoint.

To enable classifier-free guidance (CFG) and ensure graceful degradation when certain conditions are unavailable, we apply the following dropout schedule during training: text captions are replaced with empty-string T5 embeddings with a probability of $20\%$, and Observer Sink tokens are zeroed out with a probability of $20\%$. At inference we apply the same guidance scale to the actor and observer streams.

\subsubsection{Additional model details}
World Observer fine-tunes Cosmos-Predict2.5~\citep{nvidia2025worldsimulationvideofoundation}, a 2B-parameter Diffusion Transformer (DiT) with 28 blocks, 16 attention heads, and a hidden dimension of 2048. The model operates in a 16-channel latent space produced by a 3D VAE with $4\times$ temporal and $8\times$ spatial compression, so a chunk of $T{=}77{=}1{+}4\times19$ frames yields $L{=}20$ latents per stream. We render the actor at $R_\mathrm{a}{=}1280\times704$ and the observer at $R_\mathrm{o}{=}640\times320$, so each observer contributes roughly a quarter of the actor's tokens; despite its wider $360^\circ$ coverage, the observer is kept at low resolution for a compact token budget and faster training and inference, which also keeps the joint sequence tractable as observers are added.

The DiT input is assembled along the channel axis before patch embedding, concatenating the noisy target latent, the encoded warped video, a warp validity mask, a learnable view embedding, and any extra warp streams, in this order. The view embedding is a learnable $6$-channel vector with one row for perspective and one for panoramic tokens, which distinguishes the two streams. To separate multiple observers, which share the panoramic row and would otherwise be indistinguishable, we add $n\,\mathbf{e}_\text{obs}$ to the $n$-th observer of the same type, where $n$ is a $0$-based index and $\mathbf{e}_\text{obs}$ is a learnable vector. Since the first observer takes $n{=}0$, the multi-observer model reduces exactly to the single-observer one, so the single-observer checkpoint warm-starts multi-observer training without any change to the embedding table.

Camera poses are encoded separately from these channels. We build a $6$-channel Pl\"ucker ray map for each view at the latent resolution, equirectangular rays for the observer and pinhole rays for the actor, embed it with a patch embedder, and add it to the video tokens after patch embedding. This embedder is zero-initialized so that generation starts identically to the base model at the first step, keeping the warm-start clean.

The Observer Sink crops are taken at the actor resolution $R_\mathrm{a}$ and add four tokens in total, independent of the number of observers. They carry RoPE~\citep{su2023roformerenhancedtransformerrotary} positions outside the current generation window, at the offset $\Delta_\mathrm{os}{=}50$ of eq.~\ref{eq:sink_seq}, with the four cardinal crops spaced by a stride of $\delta_\mathrm{os}{=}5$.

\subsection{Real Data Processing}

\label{app:real_annotation}

\subsubsection{Panorama Video Processing} % 파노라마 Depth Estimation
\paragraph{360° video canonicalization.} Raw 360\textdegree{} footage is unstable, as the rig rotates between frames with an arbitrary vertical axis, and warping it directly would entangle camera rotation with the scene geometry we want to recover. We therefore canonicalize every clip into a rotation-stabilized, gravity-aligned ERP sequence, following 360Anything~\citep{wu2026360anythinggeometryfreeliftingimages}. We render each ERP frame to a cubemap and recover per-frame rotation with COLMAP~\citep{7780814} to cancel inter-frame spin, then estimate the gravity direction with GeoCalib~\citep{veicht2024geocaliblearningsingleimagecalibration} and apply a rotation that aligns it with the ERP down axis.

\paragrapht{Panoramic depth estimation.} Building the warping condition requires a high-resolution depth map over the full panorama. Since monocular depth estimators are trained on perspective images and cannot be applied directly to the distorted ERP panorama, we decompose the sphere into perspective views, predict depth at high resolution on each, and merge them back onto the sphere. For each depth keyframe we render 110-FOV perspective views along the vertices of an icosahedron, whose overlapping frusta cover every direction with at least two views. We pass these views to Depth Anything 3~\citep{lin2025depth3recoveringvisual} in a single batch without external extrinsics, so the model observes all views jointly, resolves their relative poses internally, and returns depth in one common scale, removing the per-view scale alignment and the drift it introduces. The per-view depths are fused into one ERP depth map by weighted averaging, and camera poses are defined from the front-view extrinsics and completed over the full trajectory by SLERP and linear interpolation. The resulting depth and poses are used for raymap conditioning and warping at training time.

\subsubsection{Text captioning} 
We generate text captions for all training clips using the Qwen3.6-27B vision-language model. Following the 4-view panoramic captioning strategy of PanoWorld~\citep{jiang2026panoworldgeometryconsistentpanoramicvideo}, we represent each panoramic frame using four $90^\circ$-FOV perspective crops corresponding to the Front, Right, Back, and Left directions, rather than directly captioning the raw equirectangular (ERP) frame. We adopt this representation to avoid directly exposing the vision-language model to the discontinuity at the ERP boundary and the strong projection distortion near the poles.

For each clip, we generate one panorama caption and one perspective caption for each camera trajectory. Each caption describes the environment, dynamic actors, and specific events in the scene. By comparing observations across multiple timestamps for each viewing direction, the caption records objects that enter, leave, or change over time, including events occurring outside the current field of view of the perspective cameras. Because these out-of-view events are recorded directly in the caption, it carries information about how the scene's dynamic state evolves outside a camera's current field of view. The panorama caption summarizes these dynamics over the full panoramic view, while each perspective caption focuses on the content visible along its corresponding camera trajectory. The panorama caption is further required to describe scene dynamics without referring to screen position or compass direction. Camera behavior is described directly in each perspective caption's prose, reflecting the sampled yaw, pitch, and field of view of the corresponding trajectory. We apply the same captioning procedure to the synthetic data.

\subsubsection{Trajectory Construction} 
For each clip, we render six perspective-camera trajectories from the same temporal segment of a panoramic video. Each trajectory independently samples its field of view and pitch, which remain fixed throughout the clip, while roll is fixed at $0^\circ$. For all trajectories except \texttt{origin}, the reference yaw is independently chosen to be either aligned with the canonical forward direction or randomly sampled.
The six trajectories consist of two static and four moving views. The \texttt{origin} trajectory remains fixed at the canonical forward direction with zero yaw and pitch, while \texttt{front\_static} remains fixed at its sampled reference orientation. The \texttt{yaw\_right} and \texttt{yaw\_left} trajectories continuously rotate to the right and left, respectively, with a randomly sampled total rotation angle and a rotation speed that varies over time. The \texttt{out\_back\_right\_first} and \texttt{out\_back\_left\_first} trajectories first rotate in the specified direction and then reverse direction. The rotation angles before and after the reversal are sampled independently, and the reversal point is randomly selected within the clip. This construction provides diverse starting viewpoints and camera motions while keeping all six perspective views temporally aligned with the source panorama.
\subsection{Synthetic Data Processing}

\label{app:synthetic_annotation}
We use CARLA~\citep{dosovitskiy2017carla} to construct our synthetic video dataset, drawing on both the Unreal Engine 4 build (v0.9.15), from which we use \texttt{Town01}, \texttt{Town02}, \texttt{Town03}, \texttt{Town05}, and \texttt{Town10}, and the Unreal Engine 5 build (v0.10.0), from which we additionally use \texttt{Town10}. We randomly spawn vehicles and pedestrians before rendering to simulate realistic traffic, and render each clip under an independently sampled weather and lighting condition, combining clear and overcast presets with noon or sunset time of day and jittering cloudiness, sun altitude, and fog, so that appearance, illumination, and traffic all vary across the dataset. We additionally render metric depth maps and extract camera parameters per frame to support geometric referencing.

\paragrapht{Panoramic actor-observer capture.}
CARLA does not natively expose an equirectangular (ERP) camera, so we assemble each panorama from a cubemap rig of six pinhole RGB-D cameras ($90^\circ$ FOV, $960\times960$ each) captured in lock-step within a single simulation tick and stitched into a $360^\circ\times180^\circ$ ERP panorama and depth map. Each clip contains one moving actor and two static observers, all captured with such rigs at an eye height of $1.6\text{--}1.8\,$m. The actor walks along these waypoints at a per-clip speed sampled uniformly from $1.3\text{--}3.5\,$m/s. The waypoints are sampled at roughly $10\,$m spacing along each map's sidewalk lanes. The two observers are placed at the waypoints nearest to the start and end of a longer rendered sequence, so the actor's offset from each observer varies across clips. Because all three rigs advance under the same simulation clock, they witness the same vehicles and pedestrians at the same instant, giving time-aligned observations of one evolving world that real footage cannot provide. For each observer we store its $t{=}0$ panorama and depth as a forward-warp source.

\section{Evaluation Details}

\subsection{Benchmark Construction}
\label{app:bench_annotation}

The benchmark videos are constructed so that every clip poses the same well-defined challenge. The camera rotates away from its starting heading and then returns to it, causing one or more dynamic objects to leave the field of view and later re-enter. Clips are drawn from the same real and synthetic (CARLA) sources used for training. We first restrict candidate scenes to those containing fewer than 10 dynamic objects in total, so that individual objects remain visually uncluttered and unambiguous to track and describe. Within each such scene, we crop a perspective view whose camera rotates away from its initial heading and then returns to it. This controlled motion deliberately causes objects to leave the frame and later re-enter, rather than relying on incidental camera motion. We then manually curate 1-3 objects per scene (human, vehicle, or animal) that unambiguously exit through a frame edge and later re-enter, discarding candidates whose exit or re-entry is ambiguous or does not occur. For the multi-observer setting, the two observer panoramas are captured at the start and final positions of the actor trajectory, both at the first timestep.

\subsection{Evaluation Metrics}
\label{app:eval_metrics}
We evaluate generation quality along four aspects: visual and temporal fidelity with FID~\citep{heusel2018ganstrainedtimescaleupdate} and FVD~\citep{unterthiner2019accurategenerativemodelsvideo}, image quality with VBench~\citep{huang2023vbenchcomprehensivebenchmarksuite}, camera-following accuracy with RotErr and TransErr, and 3D adherence with masked PSNR and LPIPS~\citep{zhang2018unreasonableeffectivenessdeepfeatures}. Both Real-OOV-Bench and Synthetic-OOV-Bench provide ground-truth videos, and all metrics are reported separately on the two benchmarks. For panorama world models, whose output is a full panorama, we crop a perspective view using the ground-truth camera rotation and compare it against the ground-truth perspective, matching our actor output. For the reference-based metrics, we resize each model's output and the ground truth to a common resolution so they are compared under the same spatial scale. Below we detail how each metric is computed.

\paragrapht{3D adherence.}
PSNR and LPIPS are computed between the generated video and the ground-truth video. Because our model targets dynamic video generation grounded in real-world references rather than exact 4D reconstruction, dynamic objects in the generated video need not match those in the ground truth. To focus these metrics on how faithfully the model preserves static scene structure, we segment dynamic objects in both generated and ground-truth frames using SAM3~\citep{carion2026sam3segmentconcepts} with text prompts for dynamic categories (pedestrian and vehicle), and compute PSNR and LPIPS only over the static regions.

\paragrapht{Camera-following accuracy.}
We extract per-frame camera extrinsics from the generated frames using Depth Anything 3~\citep{lin2025depth3recoveringvisual}, and take the ground-truth extrinsics from the same estimator applied to the ground-truth frames. Running the same estimator on both sides removes any bias between predicted and reference poses; for the synthetic benchmark we likewise use the DA3 estimates rather than the simulator poses, so that both benchmarks are evaluated identically. We process frames in non-overlapping chunks and compute relative poses by setting the first frame as the identity. Each translation trajectory is independently normalized by its maximum translation norm to remove scale ambiguity. RotErr is the mean geodesic distance on $\mathrm{SO}(3)$ between the predicted and ground-truth relative rotations, and TransErr is the mean $\ell_2$ distance between the scale-normalized relative translations, following~\citep{li2025vmemconsistentinteractivevideo}.

\begin{figure}[t!]       
    \centering    
    \includegraphics[width=\linewidth]{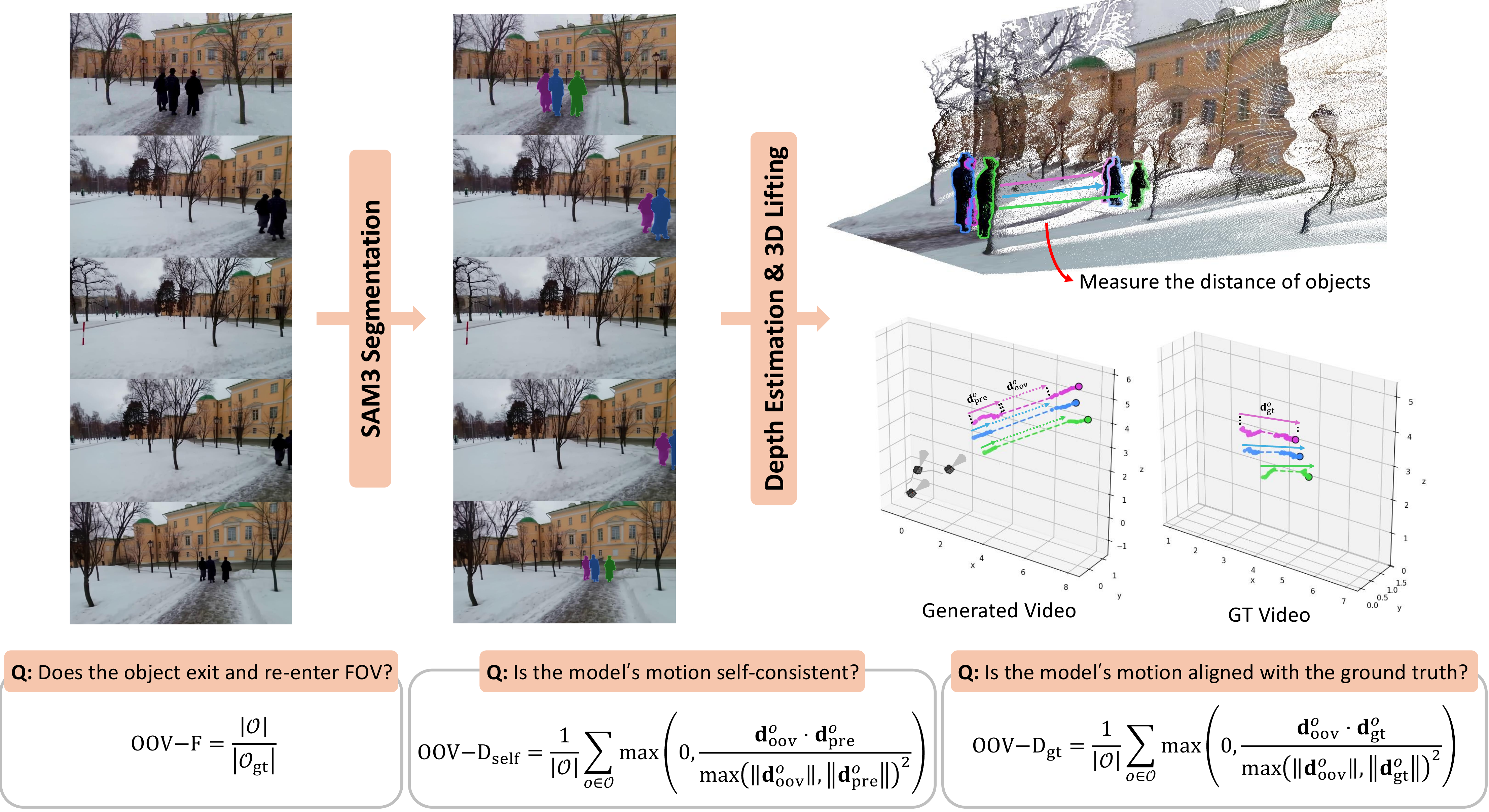}
    \vspace{-20pt}
    \caption{\textbf{Out-of-view dynamics evaluation protocol.} We track dynamic objects in 3D to evaluate their exit and re-entry (OOV-F) and motion consistency during out-of-view intervals (OOV-D$_\text{self}$ and OOV-D$_\text{gt}$).}
    \vspace{-15pt}
    \label{fig:eval_protocol}   
\end{figure}

\subsection{Out-of-View Dynamics Metrics}
\label{app:oov_metrics}
\paragrapht{Protocol.}
Let $\mathcal{O}_\text{gt}$ be the set of objects that exit and re-enter the FoV in the ground-truth video, and $\mathcal{O}\subseteq\mathcal{O}_\text{gt}$ the subset that also does so in the generated video. For each $o\in\mathcal{O}$, we localize the object with its SAM3 mask and take the median depth within the mask, which we back-project to a single 3D position, using the median to reduce sensitivity to segmentation and depth noise. The out-of-view displacement $\mathbf{d}^{o}_\text{oov}$ is then the difference between these 3D positions at the last frame before exit and the first frame after re-entry, as illustrated in Fig.~\ref{fig:eval_protocol}. Displacement is undefined outside $\mathcal{O}$, since it requires an object to actually leave and return. All methods use the same segmentation and depth pipeline.

\paragrapht{OOV-F.}
An object may fail to leave the frame as the camera moves away, or fail to return afterward, corresponding to the frame-locked and lost states, neither of which yields a well-defined displacement. We therefore report
\begin{equation}
\text{OOV-F} = \frac{|\mathcal{O}|}{|\mathcal{O}_\text{gt}|},
\end{equation}
the fraction of objects that follow the ground-truth exit-and-re-entry pattern. It is reported alongside OOV-D, since the latter is conditioned on $\mathcal{O}$ and a model that drops or locks objects is otherwise evaluated on a smaller, easier subset.

\paragrapht{OOV-D.}
We score each object's out-of-view motion by its agreement with a reference and average over $\mathcal{O}$, using the ground-truth trajectory $\mathbf{d}^{o}_\text{gt}$ (scaled to the generated out-of-view interval) and the object's own pre-exit motion $\mathbf{d}^{o}_\text{pre}$ as the two references,
\begin{equation}
\begin{aligned}
\text{OOV-D}_\text{gt} &= \frac{1}{|\mathcal{O}|}\sum_{o\in\mathcal{O}} \max\!\left(0, \frac{\mathbf{d}^{o}_\text{oov}\cdot \mathbf{d}^{o}_\text{gt}}{\max(\lVert \mathbf{d}^{o}_\text{oov}\rVert, \lVert \mathbf{d}^{o}_\text{gt}\rVert)^{2}}\right), \\[4pt]
\text{OOV-D}_\text{self} &= \frac{1}{|\mathcal{O}|}\sum_{o\in\mathcal{O}} \max\!\left(0, \frac{\mathbf{d}^{o}_\text{oov}\cdot \mathbf{d}^{o}_\text{pre}}{\max(\lVert \mathbf{d}^{o}_\text{oov}\rVert, \lVert \mathbf{d}^{o}_\text{pre}\rVert)^{2}}\right).
\end{aligned}
\end{equation}
Normalizing by the larger magnitude bounds each term in $[0,1]$ after clipping, reaching $1$ only when direction and magnitude match and $0$ for wrong-direction motion. The two references emphasize different failures. OOV-D$_\text{gt}$ penalizes any motion that departs from the ground truth, whether frozen, under- or over-shooting, or moving in the wrong direction. OOV-D$_\text{self}$ instead penalizes motion that breaks from the object's own prior trajectory, which surfaces impostor dynamics that look plausible on their own but are inconsistent with how the object was moving.

\subsection{Comparison with Existing Out-of-View Metrics}
\begin{figure}[t!]       
    \centering    
    \includegraphics[width=\linewidth]{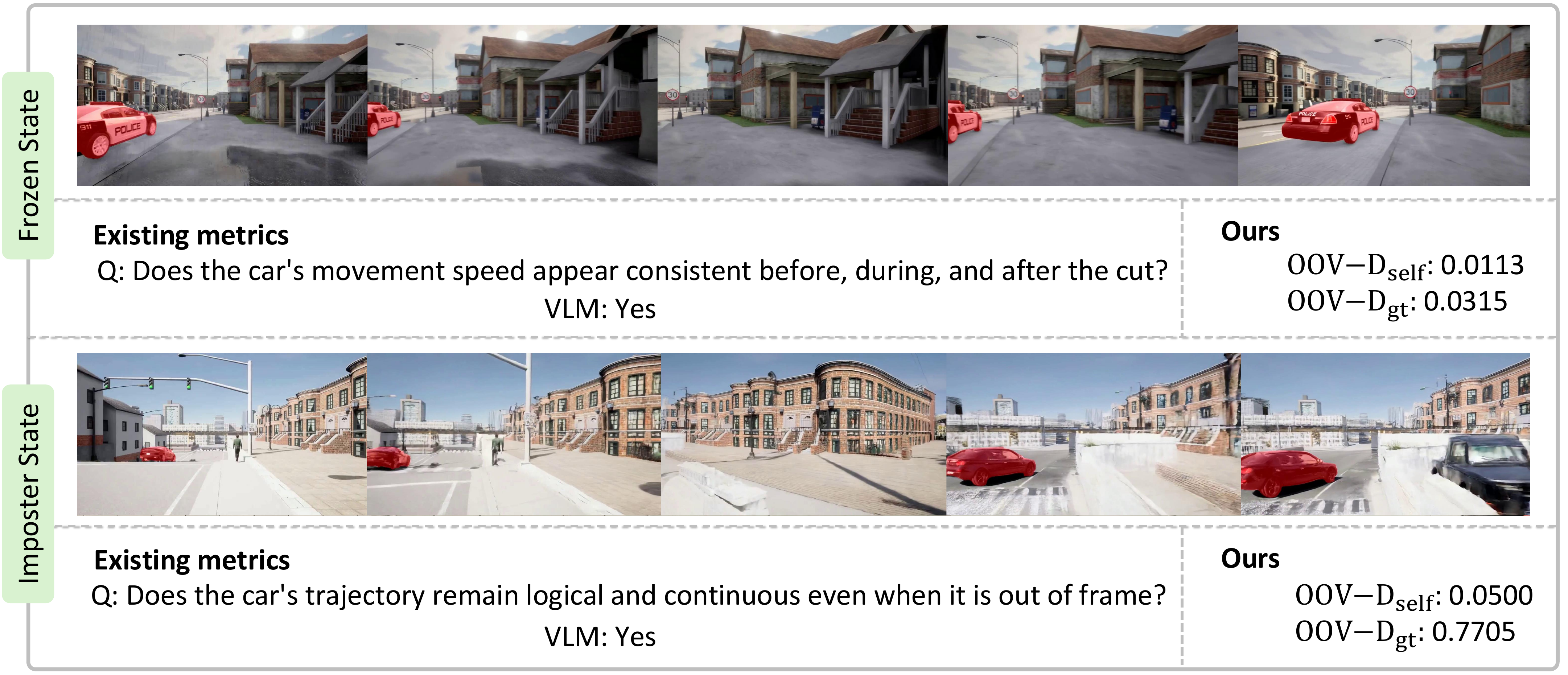}
    \vspace{-20pt}
    \caption{\textbf{Comparison with existing Out-Of-View metrics.} VLM-based metrics fail to identify frozen and impostor states as inconsistent, while our 3D displacement-based metrics quantitatively capture these failures.}
    \vspace{-15pt}
    \label{fig:metric_comparison}   
\end{figure}

Existing out-of-view benchmarks, such as MemoBench~\citep{chen2026memobenchbenchmarkingworldmodeling} and WRBench~\citep{lu2026currentworldmodelslack}, use VLM judges to decide whether an object stays consistent while out of view. The core difficulty is inherent to world models, where both the camera and the objects move, so an object's true motion cannot be read off the rendered frames directly. Judging whether a re-appearing object has evolved naturally requires recovering how it actually moved in 3D, which VLMs estimate poorly from frames alone. As a result, once an object re-appears, VLMs often cannot tell natural evolution from a frozen or impostor state, and misjudge exactly the cases these benchmarks target.

Fig.~\ref{fig:metric_comparison} shows two such cases from our evaluation. In the \textit{frozen} case, asked ``Is the object's motion consistent before, during, and after the interval?'', the VLM answers ``Yes,'' although the object has actually stopped. In the \textit{impostor} case, asked ``Does the car's trajectory remain logical and continuous even when it is out of frame?'', it again answers ``Yes,'' although the reappearing motion no longer follows the object's own prior trajectory. We follow the question format of these benchmarks, yet both judgments are wrong.

Our metric instead measures 3D displacement explicitly. We lift each object with a segmentation and depth model, so we can quantify how much it actually moved out of view. OOV-D$_\text{gt}$ exposes the frozen case, since the observed displacement is far smaller than the ground-truth reference, and OOV-D$_\text{self}$ exposes the impostor case, since the out-of-view motion departs from the object's own pre-exit trajectory. Both failures the VLM judge misses are thus identified quantitatively.

\subsection{Baseline Adaptation} 
\label{app:baselines}
Existing world models adopt different camera-conditioning interfaces, including continuous camera poses, Plücker ray embeddings, and discretized action controls. We therefore convert each benchmark camera trajectory into the representation expected by the corresponding model while preserving the underlying motion.

\paragrapht{OmniRoam~\citep{liu2026omniroamworldwanderinglonghorizon}.}
OmniRoam uses a reference-space camera trajectory defined at the latent temporal resolution. Since its released panoramic conditioning path models camera translation without explicit rotational control, we re-base the benchmark trajectory to the first frame, retain only the translational component, and map it to OmniRoam's reference coordinate system.

\paragrapht{PanoWorld~\citep{li2026panoworldrealworldpanoramicgeneration}.}
PanoWorld conditions its Action-Model on relative camera translation and yaw. We compute camera centers from the benchmark extrinsics, express their displacement in the first-camera coordinate system, and map the resulting motion to PanoWorld's panoramic coordinate convention. Relative yaw is measured with respect to the initial camera orientation, while translation is preserved at its original metric scale without additional normalization.

\paragrapht{HyDRA~\citep{chen2026sightmindhybridmemory}.}
HyDRA is evaluated using its native out-of-view reconstruction protocol rather than the shared start-frame generation protocol. For each sequence, we identify the trajectory apex and split the ground-truth video into a context segment before the apex and a target segment after it. Camera poses are then expressed relative to the first target frame and sampled at the model's latent temporal resolution. Since the model generates a fixed-length sequence, we evaluate only the frames corresponding to the valid target interval.

\paragrapht{HY-World 1.5~\citep{hyworld2025}.}
HY-World 1.5 conditions generation on both continuous camera poses through PRoPE and discretized translation and rotation actions. We subsample the benchmark trajectory to the model's latent temporal resolution, express the poses relative to the first frame, and derive the corresponding discrete motion labels from the same relative trajectory. Pose normalization follows the convention used during training.

\paragrapht{Dream-X~\citep{dreamxteam2026dreamxworld10generalpurposeinteractive}.}
Dream-X conditions generation on camera poses through Projective Rotary Position Embedding (PRoPE). We directly convert the benchmark world-to-camera trajectories into the model's RealEstate10K-style pose representation, express them relative to the first frame, and preserve the original metric translation scale.

\paragrapht{FantasyWorld~\citep{dai2025fantasyworldgeometryconsistentworldmodeling}.}
FantasyWorld represents camera motion using per-pixel 6-channel Plücker ray embeddings. We rescale the benchmark intrinsics to the generation resolution and construct the camera conditioning using the model's official Plücker embedding implementation. Following the released inference procedure, camera translations are normalized using a scene scale estimated from monocular depth of the initial frame.

\paragrapht{LingBot-World~\citep{robbyantteam2026advancingopensourceworldmodels}.}
LingBot-World conditions each transformer block on 6-channel Plücker ray maps. We convert the benchmark world-to-camera poses to camera-to-world poses and provide intrinsics that preserve the benchmark field of view in the model's reference coordinate system. Camera translation magnitude is preserved, while trajectory normalization is handled by the model's native inference pipeline.

\paragrapht{Matrix-GAME 3.0~\citep{wang2026matrixgame30realtimestreaming}.}
Matrix-GAME 3.0 conditions generation on interactive camera controls. We convert each benchmark camera trajectory into the corresponding control sequence by decomposing the relative camera motion into translational and rotational components and mapping them to the model's supported controls. The resulting sequence is applied throughout generation using the model's native inference pipeline, while preserving the direction and temporal progression of the benchmark trajectory.

\section{Additional Results and Analyses}
\subsection{Additional Qualitative Results}
\begin{figure}[t!]       
    \centering    
    \includegraphics[width=\linewidth]{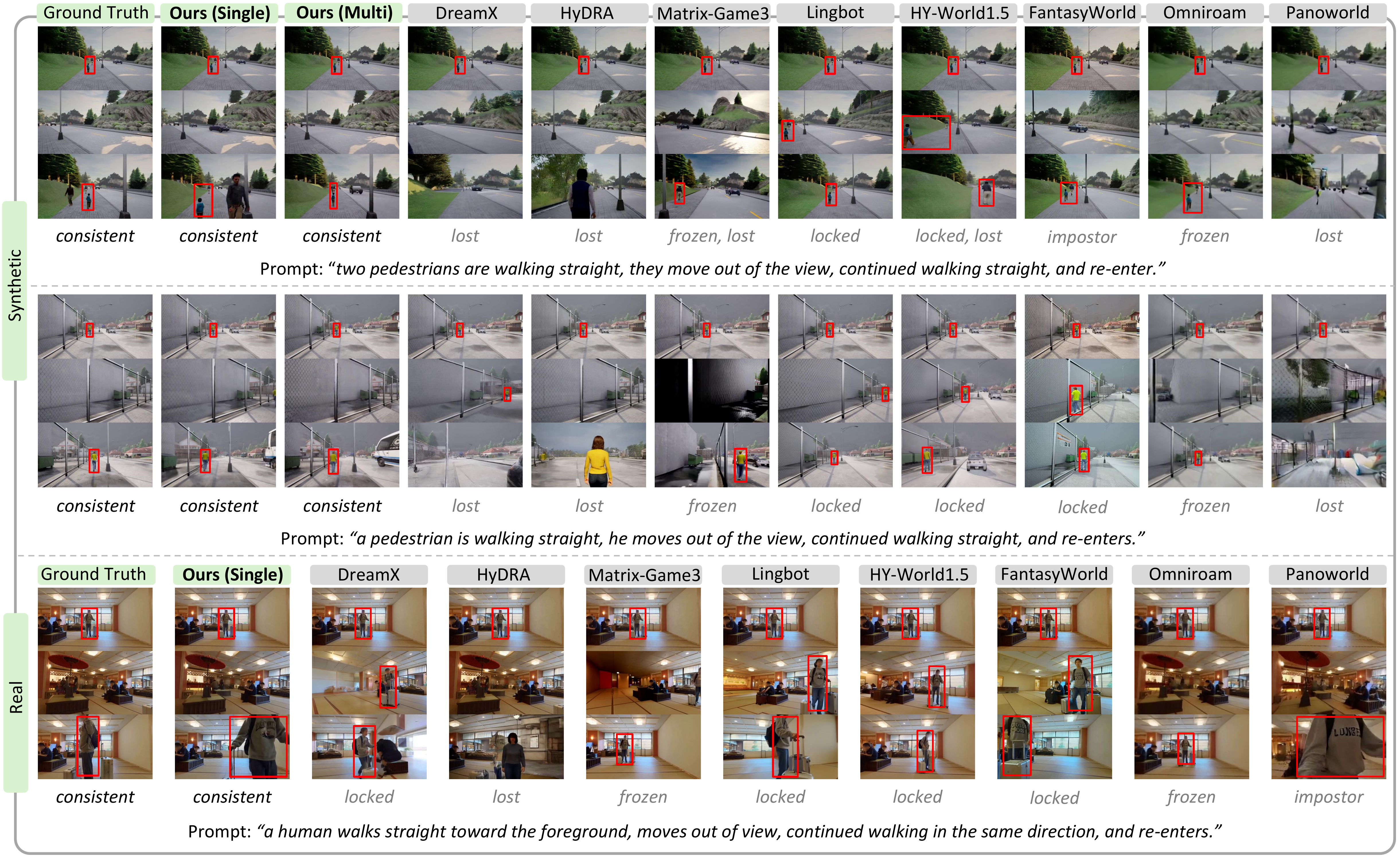}
    \vspace{-20pt}
    \caption{\textbf{Additional qualitative comparisons with other models.} We provide additional comparisons of World Observer with existing world models on real and synthetic scenes involving out-of-view dynamics. For synthetic scenes, we additionally include results from World Observer (Multi). Red boxes highlight the target objects before exit and after re-entry. Across diverse objects, motions, and scenes, existing models exhibit out-of-view failures such as lost, frozen, and impostor states, whereas World Observer consistently preserves object identity and state evolution throughout the out-of-view interval.}
    \vspace{-15pt}            
    \label{fig:additional_qualitative_comparison}   
\end{figure}

Fig.~\ref{fig:additional_qualitative_comparison} provides additional comparisons on both synthetic and real-world scenes with diverse out-of-view motions.
Existing models often fail to preserve object identity or motion once the target leaves the actor's view, leading to inconsistent re-entry.
In contrast, World Observer maintains the target state throughout the unseen interval in both single- and multi-observer settings.
The multi-observer variant further demonstrates consistent generation as the actor moves across regions covered by different observers.
Video results are available on the project page: \url{https://cvlab-kaist.github.io/world-observer}.

\subsection{Additional Qualitative Ablation Results}
\label{app:qual_ablation}
\begin{figure}[t!]       
    \centering    
    \includegraphics[width=\linewidth]{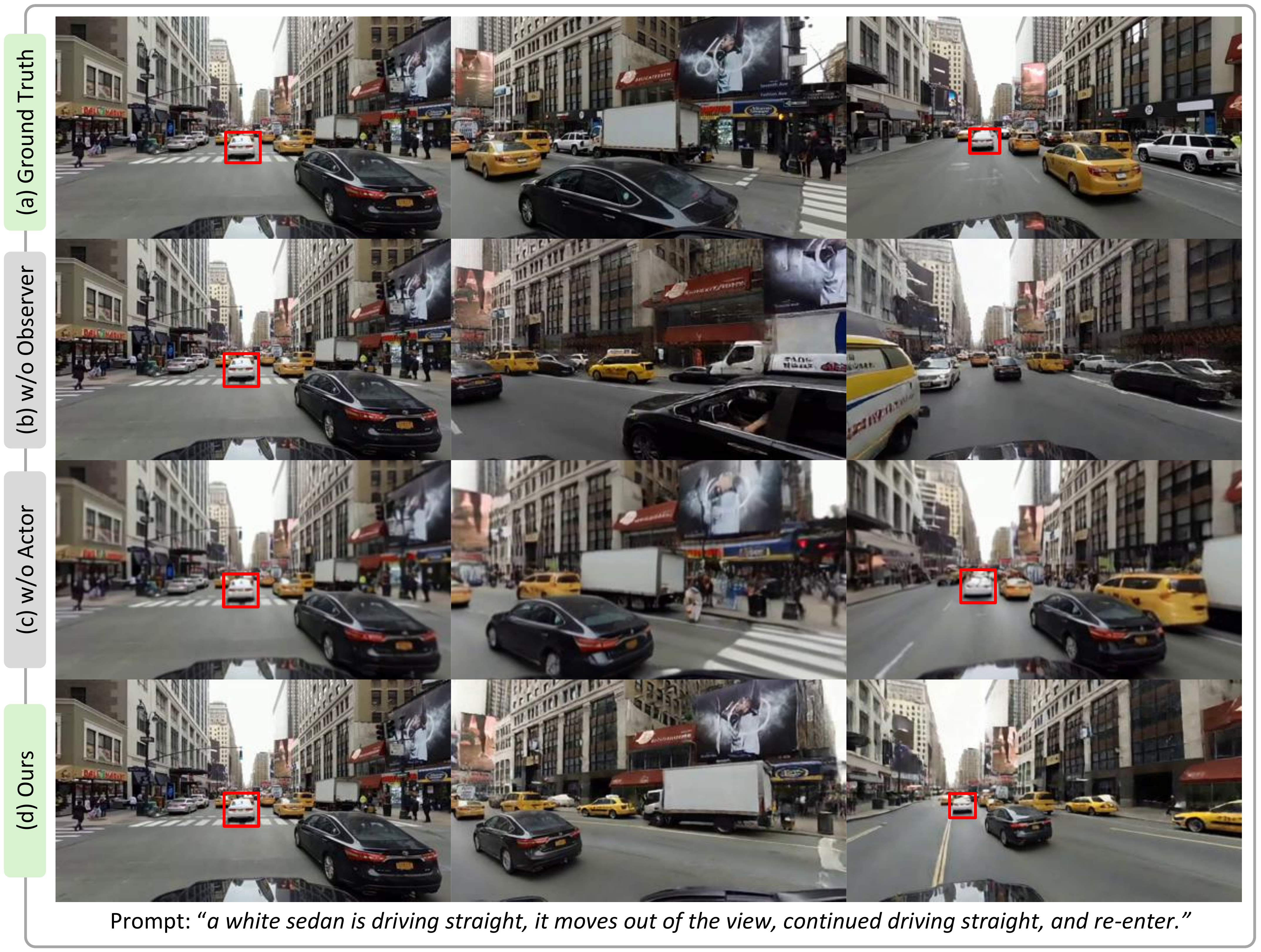}
    \vspace{-20pt}
\caption{
\textbf{Qualitative ablation results.}
We show the effect of removing each component on generation quality. Without the observer, the model fails to maintain out-of-view dynamics, causing the dynamic object to disappear when it re-enters the view. Without the actor, panorama-based generation followed by cropping leads to reduced visual fidelity. World Observer preserves evolving out-of-view dynamics while maintaining high-fidelity generation.
} 
    \vspace{-15pt}            
    \label{fig:additional_qualitative_ablation}   
\end{figure}

\begin{figure}[t!]       
    \centering    
    \includegraphics[width=\linewidth]{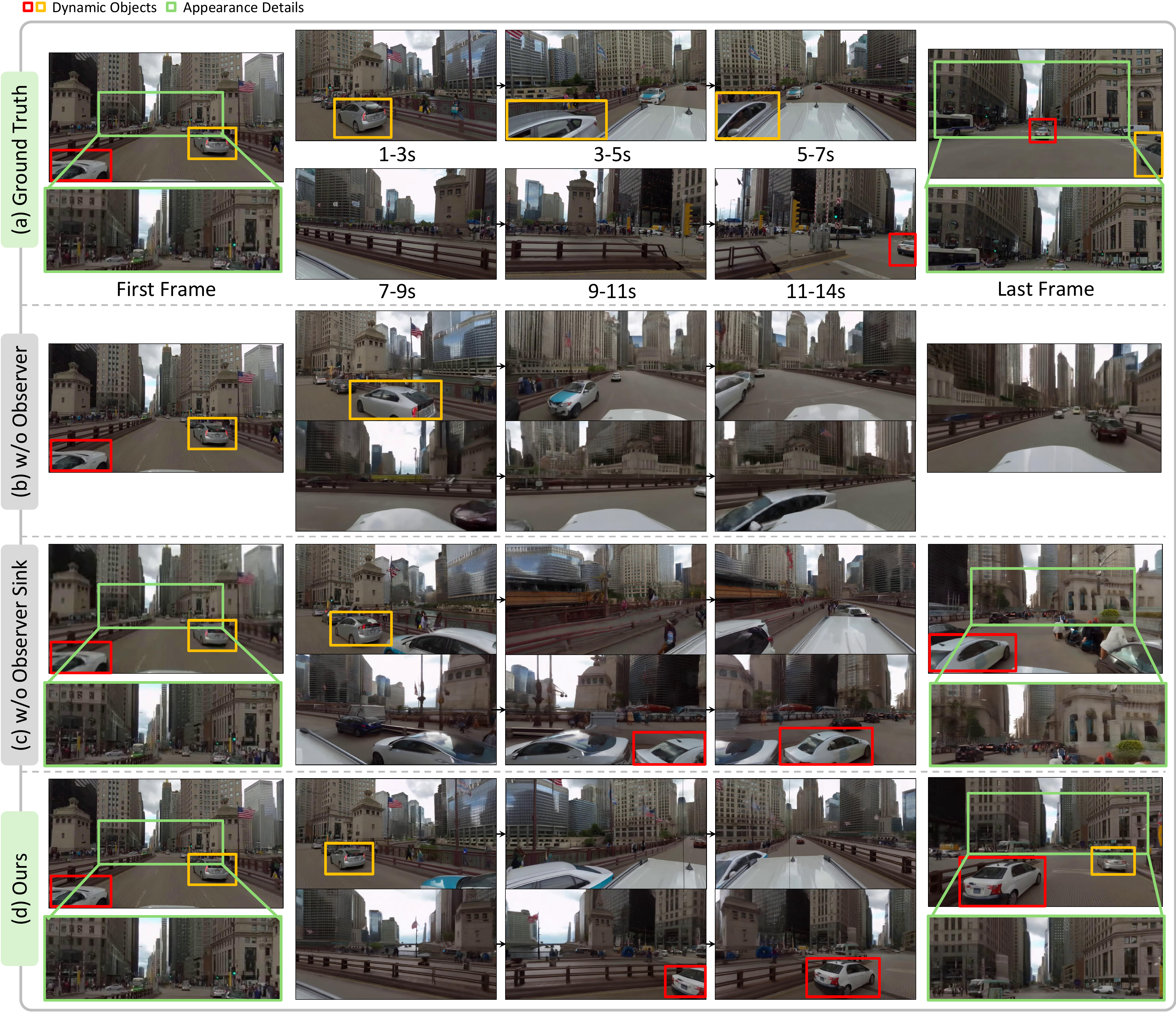}
    \vspace{-20pt}
\caption{
\textbf{Qualitative ablation results in AR generation.}
We compare ground truth with variants without the observer and Observer Sink under a long camera trajectory with a full $360^\circ$ sweep.
Without the observer, out-of-view object dynamics are not maintained, while removing the Observer Sink degrades fine appearance under large viewpoint changes.
Our full model preserves both dynamic objects (red/yellow) and appearance details (green) upon re-entry.
} 
    \vspace{-15pt}            
    \label{fig:qualitative_ablation}   
\end{figure}

\begin{figure}[t!]       
    \centering    
    \includegraphics[width=\linewidth]{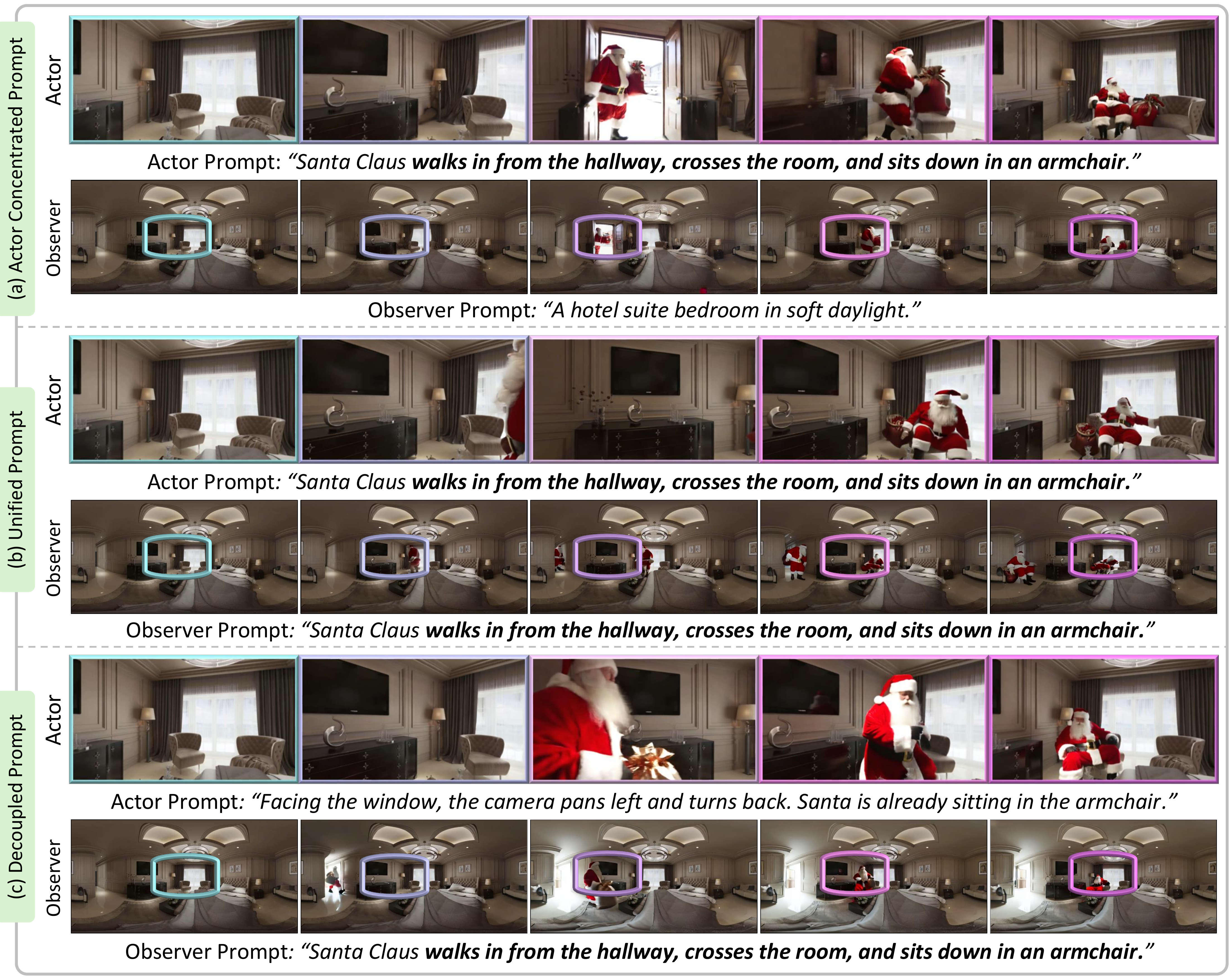}
    \vspace{-20pt}
\caption{
\textbf{Qualitative prompt ablation results.}
We compare (a) actor-concentrated, (b) unified, and (c) decoupled actor-observer prompting.
Actor-concentrated prompting misplaces off-view events in the actor view, while unified prompting can hallucinate duplicate subjects in the observer.
Decoupled prompting instead produces a consistent event across both streams.
}
    \vspace{-15pt}            
    \label{fig:qualitative_ablation_prompt}   
\end{figure}

Fig.~\ref{fig:additional_qualitative_ablation} presents qualitative ablation results on the two generation streams. Without the observer, the model cannot keep an object consistent while it is out of view, and the object re-enters with a changed state or appearance (Fig.~\ref{fig:additional_qualitative_ablation}(b)). Without the actor, generating from the panoramic observer alone yields low fidelity and visible distortion (Fig.~\ref{fig:additional_qualitative_ablation}(c)). We further analyze joint generation and the Observer Sink under autoregressive rollout (Fig.~\ref{fig:qualitative_ablation}), and the role of decoupled prompting (Fig.~\ref{fig:qualitative_ablation_prompt}).

\paragrapht{Effect of joint generation.}
The state-memory role of joint generation is most pronounced under autoregressive rollout. Without joint generation, the only information available to the actor is the Observer Sink and the recent actor history, so the model cannot maintain how the state evolves outside the field of view, and all memory is ultimately confined to the actor view. Joint generation instead lets the actor generate the dynamics within its view while tracking out-of-view dynamics through the observer. Retrieving past frames during rollout, as prior models do, only recalls observations from specific past timesteps. The model must then generate dynamics without knowing how the state progressed out of view between those observations and the present. Consistent with this, removing joint generation fails to keep out-of-view motion self-consistent and markedly lowers OOV-D$_\text{self}$ (Tab.~\ref{tab:ablation}). In contrast, from the history panorama alone, our model goes beyond static memory and continuously evolves the out-of-view dynamics that retrieval-based approaches cannot carry forward (Fig.~\ref{fig:qualitative_ablation}(b)).

\paragrapht{Effect of the Observer Sink.}
The effect of the Observer Sink becomes more visible as the camera moves more aggressively. To show this, we run inference on a trajectory that moves forward while continuously rotating to the right, sweeping a full $360^\circ$. Without the Observer Sink, so fine appearance quality degrades as the rotation grows and errors accumulate over the rollout. The Observer Sink supplies high-resolution perspective references covering the surroundings of the initial panorama, so that regions undergoing such large viewpoint changes are still rendered faithfully when they re-enter the actor view (Fig.~\ref{fig:qualitative_ablation}(c)).

\paragraph{Effect of decoupled Actor-Observer prompting.}
The world perceived by the actor and the world perceived by the observer are inevitably different. Prior world models can only be controlled through the actor prompt, so a prompt describing something outside the actor's view is forced to hallucinate within the region the actor currently sees. We instead decouple the observer prompt so that control acts on the broader world the observer covers. Fig.~\ref{fig:qualitative_ablation_prompt}(a) shows an event that can only be expressed with knowledge of the full surrounding scene. Placed in the actor prompt, it is hallucinated inside the actor's view. Placed in the observer prompt, the observer generates it in the surrounding world, and the actor observes it only when it comes into view. Fig.~\ref{fig:qualitative_ablation_prompt}(b) shows a different failure of a unified prompt. The actor, which cannot see the surroundings, hallucinates the subject in its own view, and because the two streams are generated jointly, this hallucination propagates to the observer even though the observer already places the subject correctly. The result is a duplicate, with an additional Santa appearing while another is already seated in the armchair. Decoupled prompting avoids this by assigning each stream a prompt consistent with what it should observe.

\section{Discussions}
\subsection{Limitations}
World Observer currently assumes panoramic observations as conditioning inputs, which restricts the formulation to settings where such views are available. This requirement could be relaxed by first outpainting a perspective observation into a panorama, allowing the same framework to operate when only perspective inputs are given. The same approach can be applied to multi-observer settings, where multiple panoramic conditions could be constructed from perspective observations at different locations.
Because the observer budget is limited, observing the entire world remains difficult. Nevertheless, flexible observer placement allows this limited budget to be allocated to regions of interest, preserving their dynamics independently of where the actor moves.

\subsection{Societal impact}
Decoupling observing from acting enables world models to maintain the evolving state of selected regions independently of where the actor moves or looks. This capability can benefit embodied navigation, interactive simulation, and planning, where relevant parts of the world may need to remain continuously represented even when they are not currently observed by the acting agent. Flexible and multi-observer configurations further allow a limited observation budget to be allocated to regions of interest, supporting more persistent world-state maintenance without requiring the actor itself to revisit those regions.

\end{document}